%% file: iclr2027_conference.tex
\documentclass{article} 
\usepackage{iclr2027_conference,times}

\input{math_commands.tex}

\usepackage{hyperref}
\usepackage{url}
\usepackage{graphicx}
\usepackage{caption}
\usepackage{fontawesome5} 
\usepackage{booktabs}
\usepackage{float}
\usepackage{xcolor}

\title{Closing the Loop: Practical Training Recipes for Looped Language Models}

\author{
\makebox[\textwidth][c]{%
\bfseries
\begin{tabular}{@{}ccc@{}}
Andrei Marchenko\thanks{Corresponding authors.} &
Viacheslav Bezrukov\footnotemark[1] &
Oleg Kashurin
\end{tabular}
}
\\[-3pt]
\makebox[\textwidth][c]{%
\bfseries
\begin{tabular}{@{}ccc@{}}
Inessa Fedorova &
Dmitry Bocharov &
Yuliana Shakhvalieva
\end{tabular}
}
\\[-3pt]
\makebox[\textwidth][c]{%
\bfseries
\begin{tabular}{@{}cc@{}}
Maria Tikhonova &
Valerii Ternovskii
\end{tabular}
}
\\[4pt]
\makebox[\textwidth][c]{RND NLP, DAIMLD, Russian Federation}
\\[-1pt]
\makebox[\textwidth][c]{%
\texttt{\{ammarchenko,vibezrukov\}@daimld.tech}
}
\\[4pt]
\makebox[\textwidth][c]{%
  \begingroup\small\hypersetup{hidelinks}%
  \mbox{%
    \href{https://github.com/ai-forever/OrbitLM}{\faGithub\hspace{0.35em}\textbf{Code:} OrbitLM}%
    \qquad
    \href{https://huggingface.co/ai-forever/OrbitLM-1.8B-Thinking}{%
      \raisebox{-0.20\height}{\includegraphics[height=1.10em]{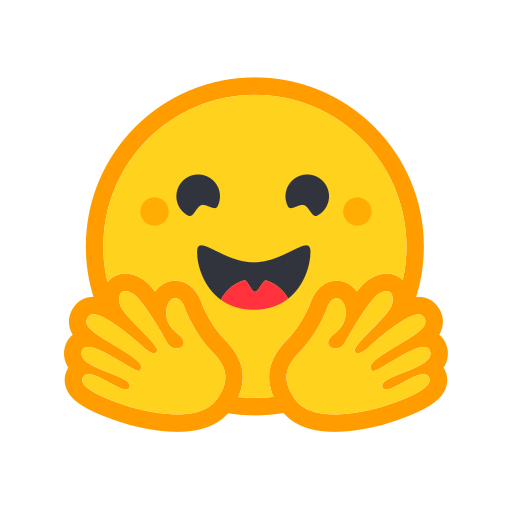}}%
      \hspace{0.35em}\textbf{Model:} OrbitLM-1.8B-Thinking}%
  }%
  \endgroup
}
}
\newcommand{\scaled}{OrbitLM}
\newcommand{\rev}[1]{#1}
\newcommand{\revcolor}{}

\iclrfinalcopy 

\begin{document}

\maketitle

\begin{abstract}
Looped language models increase effective depth by repeatedly applying a shared block of layers, but existing large-scale training recipes require multiple stages and trillions of tokens. Moreover, the benefits of recurrence remain difficult to isolate from differences in training. We address these challenges through four contributions. (1) \rev{We develop a two-stage from-scratch recipe using 310B tokens, compared to Ouro's 7.7T.} Learning-rate warmup and stronger exit-gate regularization enable stable training without staged pretraining, \rev{with stability further validated over 1.48T tokens using different data and a tokenizer.} (2) \rev{Our 1.4B LoopLM outperforms a parameter-matched dense model trained on the same data on all 12 benchmarks}, gaining 14 points on GSM8K, 10 on MATH, and 22 on DROP. At matched inference compute, \rev{it approaches a 3.9B dense model on reasoning and reading comprehension with only 37\% of its parameters, although the dense model is 1.4 times cheaper to train.} (3) We convert pretrained dense models into looped ones using a single learned input-mixing scalar and a smoothed exit loss, without step-specific parameters. \rev{Applied to Qwen3-1.7B-Base, this recipe improves over a dense baseline continued on the same data on most benchmarks across two data mixtures, with gains exceeding two standard errors on GSM8K, MATH, and MMLU-Pro on the curated mixture.} (4) \rev{LoopLM matches the decoding speed of a dense model at equal inference compute while storing fewer weights. Self-speculative decoding further provides up to a 1.65$\times$ speedup.} Together, these results \rev{establish practical recipes for looped language models and quantify the benefits and trade-offs of recurrence against dense baselines trained on the same data.}
\end{abstract}

\begin{figure}[!ht]
    \centering
    \includegraphics[width=\linewidth,height=0.22\textheight,keepaspectratio]{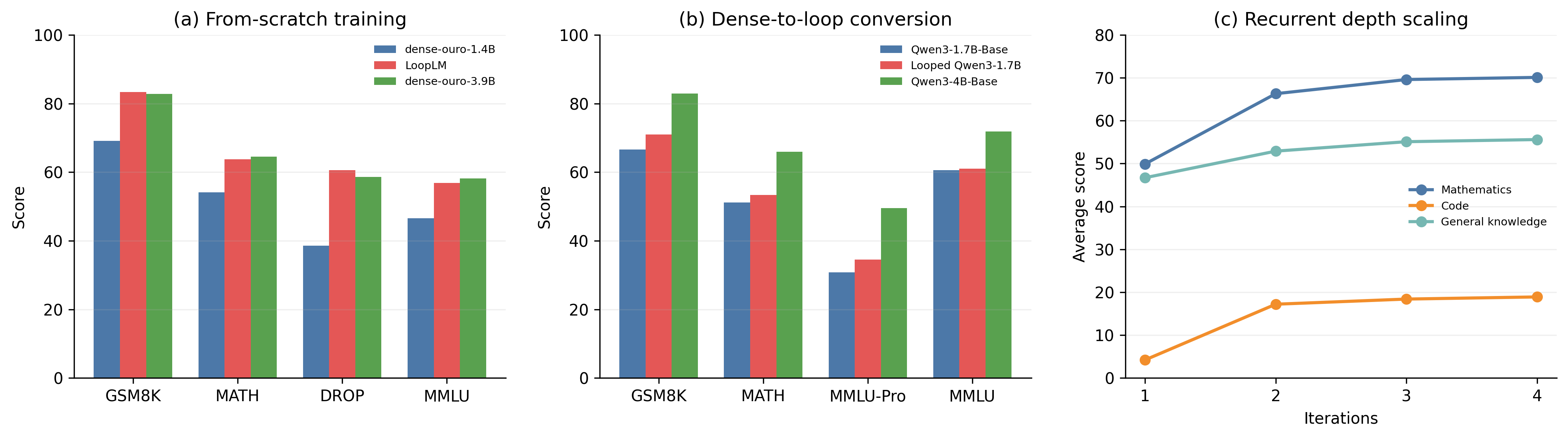}
    \captionsetup{font=footnotesize,skip=3pt}
    \caption{
    \rev{Looped models versus dense baselines trained on the same data.
    (a) Four-step LoopLM trained on 310B tokens versus parameter-matched (1.4B) and inference-compute-matched (3.9B, at three steps) dense models, evaluated on four benchmarks from Table~\ref{tab:looplm-benches-anneal}.
    (b) Looped Qwen versus Qwen3-1.7B-Base and Qwen3-4B-Base, all continued on the same data (Table~\ref{tab:looped-dense-instruct}).}
    (c) Mean scores across recurrent steps for mathematics (GSM8K, MATH, MATH500), code execution (CRUXEval), and knowledge and commonsense (six multiple-choice benchmarks from Table~\ref{tab:looplm-benches-anneal}).
    }
    \label{fig:looplm-main-results}
\end{figure}

\section{Introduction}

Transformers allocate the same amount of computation to every token, regardless of its difficulty. Looped language models offer a more flexible alternative: they repeatedly apply a shared transformer block to increase effective depth without increasing the number of parameters. With adaptive exits, they can also allocate more computation to tokens that require it \citep{dehghani2019universal,zhu2025ouro}.

Ouro demonstrated that looped language models can be pretrained at scale and compete with much larger dense models \citep{zhu2025ouro}. However, several challenges remain. First, its training pipeline involves five stages and 7.7T tokens, making reproduction costly. Second, it requires training from scratch rather than reusing existing pretrained checkpoints. Third, comparisons with dense models trained on different data make it difficult to isolate the benefits of recurrence.

In this work, we address these challenges by developing a simpler from-scratch training recipe, introducing a lightweight method for converting pretrained dense models into looped ones, and evaluating both approaches against dense baselines trained on the same data.

For training from scratch, we reduce Ouro's five-stage pipeline to two stages: pretraining followed by high-quality mid-training. Our 1.4B LoopLM is trained on 310B tokens, 25 times fewer than Ouro-1.4B, using approximately 3\% of its training compute (Appendix~\ref{app:compute}). \rev{Although it trails Ouro-1.4B on most benchmarks (Table~\ref{tab:looplm-external}), it achieves strong results for its training budget.} We also identify a key source of instability: early collapse of the exit gate. Learning-rate warmup prevents this collapse when combined with stronger KL regularization, which also allows the gate to recover if collapse occurs. The resulting LoopLM outperforms its parameter-matched dense baseline on all 12 benchmarks in Table~\ref{tab:looplm-benches-anneal}, with particularly large gains on DROP, CRUXEval-I, and GSM8K. At matched inference compute, it approaches a 3.9B dense model on mathematical reasoning and reading comprehension, although the larger model remains stronger on knowledge and commonsense tasks.

For pretrained models, we find that naively applying the Qwen3-1.7B-Base decoder recurrently degrades performance at later iterations \citep{yang2025qwen3}. We address this with a single learned scalar that mixes the original token embeddings into subsequent iterations and a smoothed exit loss that maintains training signals across all steps. This simple modification eliminates the need for step-specific adapters. Across two data mixtures, the resulting Looped Qwen outperforms Qwen3-1.7B-Base continued on the same data on most benchmarks. On the curated mixture, improvements on GSM8K, MATH, and MMLU-Pro exceed two standard errors. The larger Qwen3-4B-Base remains stronger at comparable inference compute.

Our main contributions are:
\begin{itemize}
    \item \textbf{Simplified from-scratch training.} We develop a two-stage recipe that avoids early exit-gate collapse through learning-rate warmup and stronger KL regularization. We further validate its stability on a larger run with a different tokenizer and data mixture (Section~\ref{sec:scaled}).

    \item \textbf{Minimal dense-to-loop conversion.} We show that pretrained dense models can be converted into looped models using a single learned input-mixing scalar and a smoothed exit loss, without step-specific parameters.

    \item \textbf{Controlled evaluation of recurrence.} We compare looped models with dense baselines trained on the same data and token budgets, matching parameter counts or inference compute where applicable. This allows us to distinguish the benefits of recurrence from differences in training.

    \item \textbf{Inference efficiency.} We show that LoopLM matches the decoding speed of a dense model at equal inference compute while storing fewer weights. We also demonstrate self-speculative decoding using an earlier iteration as a draft model and investigate the limits of KV-cache sharing across iterations (Section~\ref{sec:inference}).
\end{itemize}
\section{Related work}
\label{sec:related}

\paragraph{Looped and recurrent language models.}
Universal Transformers introduced the construction behind most looped models, a shared block applied repeatedly with adaptive halting \citep{dehghani2019universal,graves2016act}; fixed-point formulations replace the halting with a convergence test on the hidden state \citep{movahedi2026fixedpoint}. Looped Transformers use the shared block to extend depth and length generalization \citep{saunshi2025latent,fan2025looped}, recurrent-depth models gain from more test-time iterations \citep{geiping2025scaling}, and Mixture-of-Recursions gives each token its own depth \citep{bae2025mixture}. At scale, Ouro pretrains looped models that match much larger dense models \citep{zhu2025ouro}, HRM-Text applies hierarchical recurrence to open-domain text \citep{wang2026hrmtext}, and Loop the Loopies loops a mixture-of-experts model \citep{gao2026loopies}.

\paragraph{Training stability.}
Each of these works hits an instability of its own. Ouro saw loss spikes at eight recurrent steps and cut the depth to four and halved the KL coefficient; the first large run of \citet{geiping2025scaling} stalled with hidden states that became the same for every token; HRM-Text adds a normalization scheme and warms up its backward horizon; Mixture-of-Recursions needs balancing losses to keep its router alive. None gives a from-scratch recipe \rev{that keeps the exit gate alive without staged pretraining} or names the ingredient that \rev{does} (Section~\ref{sec:looplm-training-recipe}).

\paragraph{Comparing looped and dense models.}
Comparisons against dense models trained on the same data exist: \citet{saunshi2025latent} and Mixture-of-Recursions train parameter-matched and compute-matched baselines, HRM-Text matches training FLOPs, Loop the Loopies matches pretraining cost, and Ouro's own small-scale study finds a deeper dense model ahead of the looped one at matched forward FLOPs. Ouro and the recurrent-depth model report their headline numbers against published dense models trained on other data, so the share of the gain that comes from recurrence stays open. We measure the model we report against dense twins in both regimes, and \rev{Looped Qwen, our converted model, against its source model and Qwen3-4B-Base}. \rev{Concurrent work mostly finds looped models behind dense ones at matched compute \citep{you2026tah2,wang2026smelt,sukthanker2026nofreeloop}, and \citet{chen2026uploop} upcycle Qwen3-4B-Base with gated loops.}

\paragraph{Converting pretrained models.}
Relaxed Recursive Transformers tie the layers of a pretrained model into looping blocks with one LoRA adapter per iteration and distill from the original \citep{bae2025relaxed}; Think-at-Hard starts from Qwen3-Base and adds depth-specific adapters and a router for hard tokens \citep{fu2025thinkhard}. Both rely on step-specific adapters. We show that looping the pretrained decoder as it is degrades the representation, that one learned scalar stops this degradation and that adapters are optional.

\paragraph{Reinforcement learning.}
GRPO and DAPO \citep{shao2024deepseekmath,yu2025dapo} assume a fixed computation path; RLTT distributes credit across recurrent iterations \citep{williams2026rltt}. Ouro reports no gain from RL on its final model. A separate study asks when RL adds to a looped model (Appendix~\ref{app:rl}).

\section{A simpler from-scratch recipe}
\label{sec:looplm-training-recipe}

\subsection{Ouro's staged training}

Ouro \citep{zhu2025ouro} trains LoopLMs through five stages: pretraining at eight recurrent steps, continued pretraining at four, high-quality annealing, context extension, and supervised mid-training.
Its exit gate is optimized jointly with the language model using an entropy-regularized ELBO objective with KL coefficient $\beta$.

We simplify this recipe by collapsing the pretraining stages into a single run, reducing the total token budget, and training the dense baselines of Section~\ref{sec:baselines} through the same stages on the same data.

\subsection{A single pretraining stage}

Ouro's pretraining uses three regimes before fine-tuning: pretraining at $T{=}8$, continued pretraining at $T{=}4$, and annealing on a higher-quality mixture \citep{zhu2025ouro}.
We replace all three with one run from random initialization on a fixed mixture dominated by filtered web text, with mathematics and instruction-following data (Table~\ref{tab:looplm-pretrain-mix}).

As in Loop the Loopies \citep{gao2026loopies}, high-quality data is present from the start.
We keep only one mid-training stage afterwards (Section~\ref{sec:looplm-anneal}).

\subsection{Depth and stability}
\label{sec:looplm-stability}

Our single run departs from Ouro's first-stage configuration in depth and in schedule.

\begin{figure}[!htbp]
    \centering
    \includegraphics[width=0.95\linewidth]{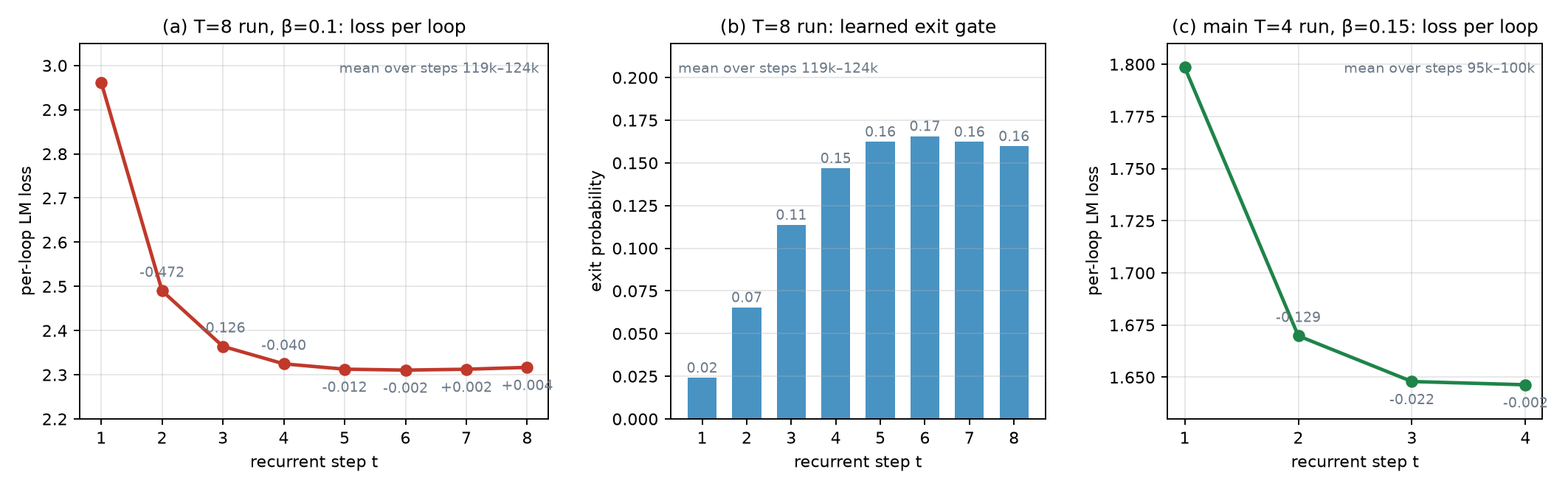}
    \caption{Why $T{=}4$. (a) Per-loop LM loss of a $T{=}8$ run ($\beta{=}0.1$, 124k steps) and the gain of each loop over the previous one: loops five to eight change the loss by about 0.01 nats or less. (b) Exit probability per loop in the same run; the gate spreads its mass evenly over loops four to eight. (c) \rev{As (a)} for the main $T{=}4$ run. Values are means over the last 5k steps.}
    \label{fig:looplm_t8_t4_comparison}
\end{figure}

Ouro explores $T{=}8$ in its first stage and drops to four afterwards, citing loss spikes at the larger depth \citep{zhu2025ouro}.
Our own $T{=}8$ run under that configuration points the same way: the per-loop loss stops improving after the third or fourth iteration, and the gate spreads its mass evenly over loops four to eight (Figure~\ref{fig:looplm_t8_t4_comparison}).
The extra iterations cost compute without measurable gain, so depth is fixed at $T{=}4$ from the start.

\begin{figure}[!htbp]
    \centering
    \includegraphics[width=\linewidth]{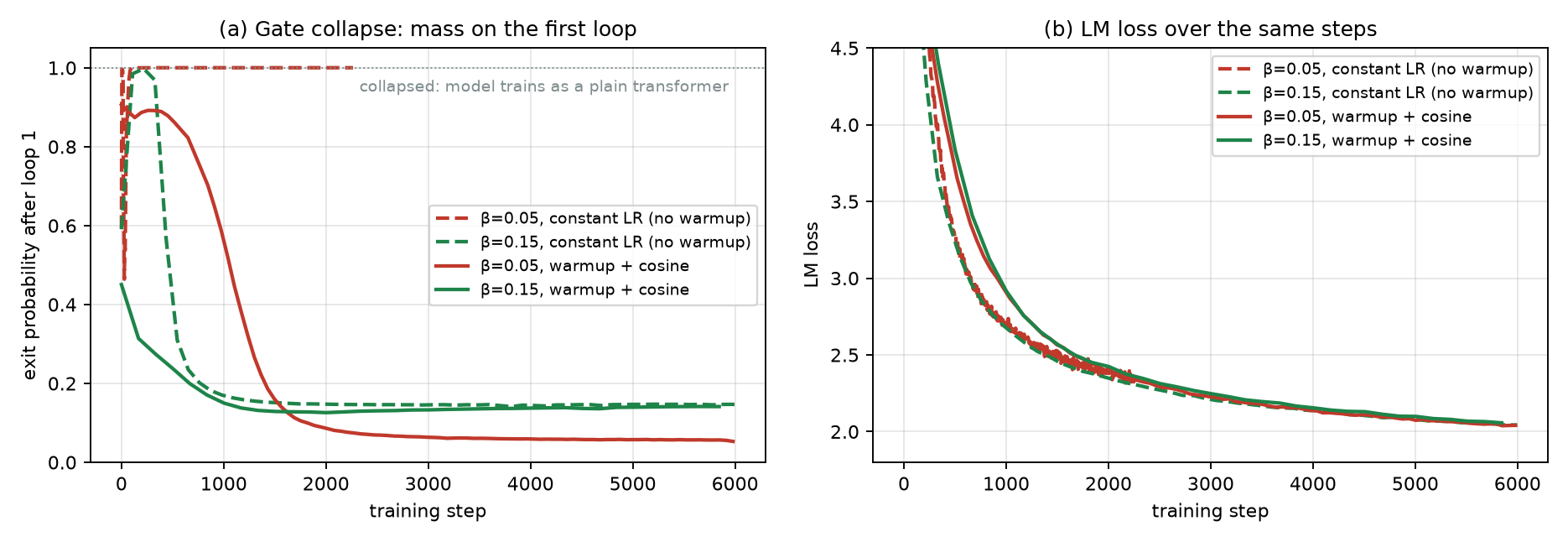}
    \caption{Gate collapse and what prevents it. (a) \rev{Exit probability $p_0$} after the first loop over the first 6k steps for $\beta \in \{0.05, 0.15\}$, with a 1k-step warmup and with Ouro's constant schedule; the constant-schedule $\beta{=}0.15$ run is separate from the cosine-schedule run of Appendix~\ref{app:early-collapse}. (b) LM loss over the same steps: the collapsed run is indistinguishable by loss alone.}
    \label{fig:looplm_stability_analysis}
\end{figure}

Ouro also lowers $\beta$ after its first stage and trains with a constant learning rate from then on.
Our run has no such transition, so we ask what keeps the exit gate alive once the schedule is gone.
The failure is a gate collapse: early in training the gate learns to exit after the first loop, later iterations stop receiving gradient, and the model trains on as a plain transformer.
Figure~\ref{fig:looplm_stability_analysis} compares $\beta{=}0.05$, the value Ouro uses after its first stage, with our $\beta{=}0.15$, each trained without warmup and with our warmup-plus-cosine schedule.
\rev{Without warmup both collapse to $p_0$ near 1: the larger $\beta$ recovers within about 1,000 steps, and the smaller one stays collapsed until its run ends at step 2,267.
With the warmup schedule the larger $\beta$ never comes near the collapsed regime, while the smaller one stays near $p_0=0.9$ for about 500 steps and recovers by about step 1,500.}
Thus, warmup prevents early collapse at our $\beta$, while larger $\beta$ enables recovery if collapse occurs: without warmup \rev{$\beta{=}0.10$ and $0.15$ recover}, and only $\beta{=}0.05$ stays collapsed \rev{for as long as we ran it} (Appendix~\ref{app:early-collapse}).
The main run therefore uses $\beta{=}0.15$ with warmup and cosine decay (Table~\ref{tab:our-recipe}).
Notably, the language-modelling loss does not reveal the collapse while it occurs, so the exit distribution must be monitored directly.

\subsection{High-quality mid-training}
\label{sec:looplm-anneal}

Despite the mathematics and instruction-following data in pretraining, the resulting checkpoint reaches only 35\% accuracy on 5-shot GSM8K (Appendix~\ref{app:rl}).
We therefore add a high-quality mid-training stage using instruction--response pairs with loss on response tokens only.
The mixture emphasizes mathematics and reasoning while retaining a knowledge-preserving component.

Mid-training continues for 100B tokens on the HRM-Text mixture \citep{wang2026hrmtext}, whose largest components are FLAN, SYNTH, and OpenMathInstruct-2 (Table~\ref{tab:looplm-anneal-mix}).
Recurrent depth, $\beta$, weight decay, optimizer, and architecture remain unchanged from pretraining.
The same stage is applied to both dense baselines with identical data, schedule, and token budget.

\subsection{Our improved setup}
\label{sec:looplm-final-setup}

The final recipe keeps Ouro's shared recurrent architecture and trains it in two stages instead of five: a single 210B-token pretraining run from random initialization followed by 100B tokens of high-quality mid-training.
The resulting 310B-token budget is 25$\times$ smaller than the 7.7T tokens used for Ouro-1.4B and needs about 3\% of the training compute of Ouro's schedule on the same architecture (Appendix~\ref{app:compute}).
Training uses $T{=}4$ from the start, a learning-rate warmup to prevent early exit-gate collapse, and $\beta{=}0.15$ to enable recovery if collapse occurs (Section~\ref{sec:looplm-stability}).
Table~\ref{tab:our-recipe} summarizes the recipe; Appendix~\ref{app:looplm-setups} provides the architecture, gate objective, and data mixtures.

\begin{table}[!htbp]
\centering
\caption{Our two-stage training recipe. Both stages share the architecture, recurrent depth, and optimizer; the main difference is the data and learning-rate schedule.}
\label{tab:our-recipe}
\small
\setlength{\tabcolsep}{4pt}
\begin{tabular}{lcc}
\toprule
 & \textbf{Pretraining} & \textbf{Mid-training} \\
\midrule
Initialization
    & From scratch
    & Pretraining checkpoint \\
Peak learning rate
    & $3\times10^{-4}$
    & $1\times10^{-4}$ \\
Final learning rate
    & $3\times10^{-5}$
    & $1\times10^{-5}$ \\
LR scheduler
    & \multicolumn{2}{c}{Warmup + cosine decay} \\
LR warmup steps
    & \multicolumn{2}{c}{1,000} \\
Weight decay
    & \multicolumn{2}{c}{0.1} \\
Gradient norm clip
    & \multicolumn{2}{c}{1.0} \\
Optimizer
    & \multicolumn{2}{c}{AdamW ($\beta_1{=}0.9,\ \beta_2{=}0.95$)} \\
Sequence length
    & \multicolumn{2}{c}{4K} \\
Training steps
    & 100,000
    & 47,684 \\
Training tokens
    & 210B
    & 100B \\
Loss
    & Next token
    & Response tokens only \\
Recurrent steps ($T$)
    & \multicolumn{2}{c}{4} \\
KL coefficient $\beta$
    & \multicolumn{2}{c}{0.15} \\
\bottomrule
\end{tabular}
\end{table}

The mid-training stage contributes most of the reasoning improvement: relative to the step-100{,}000 pretraining checkpoint, \rev{GSM8K rises from 35.3 (5-shot, plain format) to 83.3 (0-shot, the chat format of Table~\ref{tab:looplm-benches-anneal}), although the formats differ, and 5-shot MMLU improves by about nine points.}
\paragraph{\rev{Reinforcement learning.}}
\label{sec:looplm-rl}
\rev{None of the controlled comparisons in Section~\ref{sec:experiments} involves reinforcement learning. Appendix~\ref{app:rl} studies RLTT \citep{williams2026rltt} separately: it left MATH unchanged on a mid-trained checkpoint of an earlier run that already solved about three quarters of its rollouts, and added twelve GSM8K points to the pretraining checkpoint, which solved about 30\%. On the larger run of Section~\ref{sec:scaled}, RL also collapsed the exit gate onto earlier iterations when the exit probabilities received the policy gradient, a failure like that of Section~\ref{sec:looplm-stability}.}

\section{Looping pretrained LLMs}
\label{sec:looped-dense}

Only a few pretrained looped checkpoints are publicly available, making dense-to-loop conversion an attractive alternative to training from scratch.
We find that naive recurrence degrades later steps, while a single input-preserving scalar together with a smoothed exit loss is sufficient to convert Qwen3-1.7B-Base, at under 1\% of its pretraining compute (Appendix~\ref{app:compute}).
Appendix~\ref{app:looped-dense-dynamics} reports the corresponding training dynamics.

\subsection{Conversion recipe}
\label{sec:conversion-recipe}

We start from Qwen3-1.7B-Base \citep{yang2025qwen3} and apply the full decoder recurrently for $T{=}3$ steps.
Without input re-injection, later steps are consistently worse than the first: after the initial phase, the LM losses of steps 2 and 3 remain about 0.01 and 0.02 nats higher, respectively (Figure~\ref{fig:looped-dense-dynamics}a).
A learned per-step projection of the original embedding and recurrent state also fails to produce refinement: the gate abandons the middle step and the final step does not improve over the first (Figure~\ref{fig:looped-dense-dynamics}c).

We instead preserve direct access to the original embedding with a single learned scalar.
Let $e$ denote the token embedding and $h_{t-1}$ the output of the previous recurrent step.
For $t>0$,
\begin{equation}
z_t = w e + (1-w)h_{t-1}, \qquad w = \sigma(a),
\label{eq:weighted-embedding-injection}
\end{equation}
where $a$ is the only \rev{parameter the injection adds}.
With the same data, batch, and seed as the naive run, steps 2 and 3 finish about 0.02 nats below step 1 (Figure~\ref{fig:looped-dense-dynamics}b).
We initialize $w=0.99$ \rev{(0.97 in Figure~\ref{fig:looped-dense-dynamics}b)}; it \rev{changes} slowly (Figure~\ref{fig:looped-dense-dynamics}e; Appendix~\ref{app:looped-dense-setups}).

\subsection{Training stability}
\label{sec:looped-dense-stability}

The exit gate \rev{can introduce} a separate instability.
Under the standard LoopLM objective \citep{zhu2025ouro},
$\sum_t p_t\mathcal{L}_t$, each step is trained in proportion to its current exit probability; once the gate assigns little mass to a step, that step receives little learning signal.
In an unsmoothed run with rank-8 stepwise LoRA, the gate drops the middle exit and its loss rises to about 10 nats (Figure~\ref{fig:looped-dense-dynamics}d).

We therefore smooth only the exit distribution used for the LM loss:
\begin{equation}
\mathcal{L}_{\mathrm{LM}}^{\mathrm{smooth}}
  = (1-\epsilon)\sum_{t=0}^{T-1}p_t\mathcal{L}_t
  + \frac{\epsilon}{T}\sum_{t=0}^{T-1}\mathcal{L}_t.
\label{eq:exit-probability-smoothing}
\end{equation}
Gate regularization and inference still use the original exit probabilities.
With $\epsilon{=}0.05$ and $T{=}3$, every step retains at least $0.0167$ LM weight, \rev{and the step-2 loss stays low} (Figure~\ref{fig:looped-dense-dynamics}d).
\rev{Uniform weights for an initial phase followed by the gate, which corresponds to $\epsilon{=}1$ and then $\epsilon{=}0$, is an alternative we did not test.}

\subsection{Data mixes}
\label{sec:looped-dense-data}

We use a curated continued-training mixture covering mathematics, code, STEM, science, and instruction following, with final-answer supervision and no explicit chain-of-thought traces (Appendix~\ref{app:looped-dense-setups}).
\rev{To test whether the gains depend on the data}, we repeat the conversion on a second mixture dominated by filtered web text.
Each mixture is shared unchanged between the looped model and its dense baselines.

\subsection{Stepwise LoRA}
\label{sec:looped-dense-lora}

We test whether step-specific parameters are necessary by adding independent rank-32 LoRA adapters to each recurrent step \citep{bae2025relaxed}.
They provide little benefit: \rev{later-step gains nearly match those of} the shared model (Figure~\ref{fig:looped-dense-dynamics}f), and benchmark differences stay within two points with neither variant consistently ahead (Section~\ref{sec:looped-dense-results}).
We therefore use the shared model by default.

\subsection{Final architecture}

Figure~\ref{fig:looped-dense-arch} \rev{in Appendix~\ref{app:looped-dense-setups}} summarizes the final model: a shared pretrained decoder, scalar input injection from Eq.~\ref{eq:weighted-embedding-injection}, and a shared LM head and exit gate trained with Eq.~\ref{eq:exit-probability-smoothing}.
Because the third step improves training loss by less than 0.001 nats over the second on the instruction mixture (Figure~\ref{fig:looped-dense-dynamics}f), evaluation uses a fixed exit after two steps.

\section{Experiments}
\label{sec:experiments}

\subsection{Baselines}
\label{sec:baselines}

We evaluate both from-scratch and converted looped models against dense baselines
trained on the same data, stages, and token budget.
For each track, we use a \rev{baseline with the same parameters, or the source checkpoint for conversion,} and a larger
inference-compute-matched baseline.

\subsubsection{From-scratch LoopLM}

Both dense baselines remove recurrence ($T{=}1$) while retaining the same
transformer block and two-stage training procedure.
\texttt{dense-ouro-1.4B} uses one 24-layer block and matches LoopLM in parameter
count; \texttt{dense-ouro-3.9B} stacks three such blocks into 72 layers and
matches the inference compute of LoopLM at three recurrent steps.
Because LoopLM trains with all four iterations, its per-token training compute
is about $4\times$ that of \texttt{dense-ouro-1.4B} and $1.4\times$ that of
\texttt{dense-ouro-3.9B} (Appendix~\ref{app:compute}).

\subsubsection{Looped dense}

For conversion, \rev{Qwen3-1.7B-Base, the source checkpoint, is the baseline} and
Qwen3-4B-Base the inference-compute-matched reference \citep{yang2025qwen3}.
\rev{Looped Qwen trains an untied output head initialized from the embeddings,
which adds 311M trainable parameters, while the baseline keeps the tied head of
Qwen3-1.7B-Base.}
All models are continued on the same conversion mixture with identical
optimization and evaluation settings.
At its two-step exit, Looped Qwen uses \rev{22\%} fewer inference FLOPs than
Qwen3-4B-Base (Appendix~\ref{app:compute}).

\subsection{LoopLM results}
\label{sec:looplm-results}

\subsubsection{Against dense models trained on the same data}

Table~\ref{tab:looplm-benches-anneal} \rev{and Figure~\ref{fig:looplm-main-results} compare} LoopLM with both
baselines.

\begin{table}[!htbp]
\caption{Benchmark scores after mid-training (Section~\ref{sec:looplm-anneal}).
LoopLM is evaluated at a fixed exit after $k$ iterations;
\texttt{dense-ouro-3.9B} matches the inference compute of three iterations.}
\label{tab:looplm-benches-anneal}
\begin{center}
\resizebox{\linewidth}{!}{%
\begin{tabular}{lcccccc}
\toprule
Benchmark &
dense-ouro-1.4B &
\multicolumn{4}{c}{LoopLM (ours)} &
dense-ouro-3.9B \\
\cmidrule(lr){3-6}
 & & 1 iter. & 2 iter. & 3 iter. & 4 iter. & \\
\midrule
\multicolumn{7}{l}{\emph{Math and reading comprehension}} \\
GSM8K
& $69.14 \pm 1.27$
& $60.65 \pm 1.35$
& $76.50 \pm 1.17$
& $81.58 \pm 1.07$
& $\mathbf{83.32 \pm 1.03}$
& $82.79 \pm 1.04$ \\
MATH
& $54.10 \pm 0.67$
& $44.18 \pm 0.67$
& $60.90 \pm 0.65$
& $63.64 \pm 0.64$
& $63.80 \pm 0.64$
& $\mathbf{64.54 \pm 0.64}$ \\
MATH500
& $55.00 \pm 2.22$
& $45.00 \pm 2.22$
& $61.40 \pm 2.18$
& $\mathbf{63.60 \pm 2.15}$
& $63.20 \pm 2.16$
& $\mathbf{63.60 \pm 2.15}$ \\
DROP (F1)
& $38.59 \pm 0.48$
& $22.20 \pm 0.41$
& $45.02 \pm 0.49$
& $57.40 \pm 0.49$
& $\mathbf{60.57 \pm 0.48}$
& $58.57 \pm 0.49$ \\
\midrule
\multicolumn{7}{l}{\emph{Knowledge and commonsense}} \\
MMLU
& $46.52 \pm 0.41$
& $40.77 \pm 0.41$
& $52.61 \pm 0.41$
& $55.93 \pm 0.40$
& $56.84 \pm 0.40$
& $\mathbf{58.15 \pm 0.40}$ \\
MMLU-Pro
& $17.48 \pm 0.34$
& $13.80 \pm 0.31$
& $20.53 \pm 0.36$
& $24.27 \pm 0.38$
& $24.53 \pm 0.38$
& $\mathbf{28.22 \pm 0.40}$ \\
ARC-Challenge
& $46.16 \pm 1.46$
& $45.99 \pm 1.46$
& $50.60 \pm 1.46$
& $50.51 \pm 1.46$
& $51.02 \pm 1.46$
& $\mathbf{55.03 \pm 1.45}$ \\
ARC-Easy
& $74.75 \pm 0.89$
& $72.18 \pm 0.92$
& $76.47 \pm 0.87$
& $77.78 \pm 0.85$
& $77.99 \pm 0.85$
& $\mathbf{79.67 \pm 0.83}$ \\
HellaSwag
& $57.28 \pm 0.49$
& $51.02 \pm 0.50$
& $56.94 \pm 0.49$
& $58.92 \pm 0.49$
& $59.29 \pm 0.49$
& $\mathbf{67.91 \pm 0.47}$ \\
Winogrande
& $59.43 \pm 1.38$
& $56.67 \pm 1.39$
& $60.22 \pm 1.38$
& $62.90 \pm 1.36$
& $63.93 \pm 1.35$
& $\mathbf{65.19 \pm 1.34}$ \\
\midrule
\multicolumn{7}{l}{\emph{Code execution}} \\
CRUXEval-O
& $6.15 \pm 0.76$
& $6.03 \pm 0.74$
& $18.20 \pm 1.28$
& $17.36 \pm 1.23$
& $16.60 \pm 1.22$
& $\mathbf{18.48 \pm 1.28}$ \\
CRUXEval-I
& $4.61 \pm 0.60$
& $2.34 \pm 0.45$
& $16.20 \pm 1.15$
& $19.46 \pm 1.22$
& $21.21 \pm 1.29$
& $\mathbf{21.80 \pm 1.35}$ \\
\bottomrule
\end{tabular}}
\end{center}
\end{table}

At equal parameter count, four-step LoopLM outperforms
\texttt{dense-ouro-1.4B} on all 12 benchmarks.
The gains are largest on reasoning-intensive tasks: +22 points on DROP,
+17 on CRUXEval-I, +14 on GSM8K, and about +10 on MATH, MMLU, and
CRUXEval-O.
Thus, \rev{on these benchmarks, recurrence beats} applying the same
parameter budget only once.

At equal inference compute the larger dense model is ahead or level on every
benchmark, and the advantage of LoopLM at this point is its size.
With 1.4B parameters, \rev{37\%} of those of \texttt{dense-ouro-3.9B}, three-step
LoopLM comes within one standard error of the difference on GSM8K, MATH and
MATH500 and within about one point on DROP.
A fourth iteration, roughly one third more inference compute, moves it ahead
on DROP and level on GSM8K.
The gap that remains is \rev{mostly} in knowledge and commonsense benchmarks, \rev{1} to 9
points, where \rev{the dense model's extra parameters may} store more facts.

The depth sweep shows that these gains come from using recurrent computation
rather than from the underlying 1.4B checkpoint alone.
One iteration trails the parameter-matched dense model on every benchmark;
two iterations outperform it on 11 of 12, three on all 12, and the fourth
iteration improves 10 of 12 scores further.

\paragraph{\rev{Against published models.}}
\rev{Appendix~\ref{app:external} compares LoopLM with published models between 1B and 4B
parameters. Trained on 25$\times$ fewer tokens, it trails Ouro-1.4B on every benchmark
there except GSM8K, where the protocols differ. That comparison is contextual, and the
controlled evidence for recurrence remains Table~\ref{tab:looplm-benches-anneal}.}

\subsection{Looped dense results}
\label{sec:looped-dense-results}

We run the conversion twice from Qwen3-1.7B-Base using the same recipe:
once on 84B tokens of web-scale data and once on 60B tokens of curated
instruction data (Section~\ref{sec:looped-dense-data}).
Each run has dense baselines continued on the same data for the same number of
steps.
Table~\ref{tab:looped-dense-instruct} reports the instruction run; the web-scale
run is in Appendix~\ref{app:looped-dense-web}.
Looped Qwen trains with all three recurrent steps and is evaluated after two,
where its inference cost is close to Qwen3-4B-Base.

\begin{table}[!htbp]
\caption{Conversion on the curated instruction mixture, 60B tokens, evaluated
at a fixed exit after two of three recurrent steps.
$\pm$ is the standard error; $^\dagger$ denotes continued training on the same
mixture for the same number of steps; bold marks Looped Qwen ahead of
Qwen3-1.7B-Base$^\dagger$. \rev{BoolQ is 0-shot, DROP 3-shot F1 and CRUXEval-O
2-shot pass@1, evaluated on the same checkpoints.}}
\label{tab:looped-dense-instruct}
\begin{center}
\small
\begin{tabular}{lcccc}
\toprule
Benchmark &
Qwen3-1.7B-Base$^\dagger$ &
Looped Qwen, shared &
Looped Qwen, LoRA-32 &
Qwen3-4B-Base$^\dagger$ \\
\midrule
GSM8K
& $66.64 \pm 1.30$
& $\mathbf{70.96 \pm 1.25}$
& $\mathbf{69.90 \pm 1.26}$
& $82.87 \pm 1.04$ \\
MATH
& $51.14 \pm 0.66$
& $\mathbf{53.34 \pm 0.66}$
& $\mathbf{54.46 \pm 0.67}$
& $65.92 \pm 0.63$ \\
MMLU
& $60.62 \pm 0.39$
& $\mathbf{61.05 \pm 0.39}$
& $\mathbf{60.99 \pm 0.39}$
& $71.83 \pm 0.36$ \\
MMLU-Pro
& $30.76 \pm 0.41$
& $\mathbf{34.52 \pm 0.42}$
& $\mathbf{36.24 \pm 0.43}$
& $49.52 \pm 0.44$ \\
ARC-Challenge
& $53.84 \pm 1.46$
& $\mathbf{55.46 \pm 1.45}$
& $\mathbf{54.95 \pm 1.45}$
& $64.42 \pm 1.40$ \\
ARC-Easy
& $79.97 \pm 0.82$
& $\mathbf{80.13 \pm 0.82}$
& $\mathbf{80.51 \pm 0.81}$
& $86.32 \pm 0.71$ \\
HellaSwag
& $63.35 \pm 0.48$
& $\mathbf{64.13 \pm 0.48}$
& $\mathbf{64.29 \pm 0.48}$
& $70.88 \pm 0.45$ \\
Winogrande
& $62.90 \pm 1.36$
& $\mathbf{63.61 \pm 1.35}$
& $\mathbf{64.01 \pm 1.35}$
& $67.80 \pm 1.31$ \\
\rev{BoolQ}
& \rev{$80.28 \pm 0.70$}
& \rev{$79.08 \pm 0.71$}
& \rev{$80.06 \pm 0.70$}
& \rev{$84.50 \pm 0.63$} \\
\rev{DROP (F1)}
& \rev{$28.85 \pm 0.42$}
& \rev{$\mathbf{29.17 \pm 0.43}$}
& \rev{$28.07 \pm 0.42$}
& \rev{$47.91 \pm 0.49$} \\
\rev{CRUXEval-O}
& \rev{$30.48 \pm 1.56$}
& \rev{$29.90 \pm 1.53$}
& \rev{$28.58 \pm 1.51$}
& \rev{$40.64 \pm 1.66$} \\
\bottomrule
\end{tabular}
\end{center}
\end{table}

The conversion \rev{helps in} both data regimes.
Without step-specific adapters, Looped Qwen improves over
Qwen3-1.7B-Base continued on exactly the same data for the same number of
steps on \rev{all eight benchmarks of the web-scale run and on nine of the
eleven in Table~\ref{tab:looped-dense-instruct}; on BoolQ and CRUXEval-O it is
below the baseline, by less than two standard errors of the difference.}
Most gains are modest, but the curated instruction mixture produces clear
reasoning improvements: +4.3 points on GSM8K, +3.8 on MMLU-Pro, and +2.2 on
MATH, each exceeding two standard errors of the difference.
On the web-scale mixture, gains remain within about two points, \rev{and} only MMLU
and HellaSwag \rev{exceed two standard errors}.

The larger gains on the curated mixture suggest that the benefit
depends strongly on the continued-training data.
The effect does not require step-specific capacity: the shared and
LoRA-32 variants differ by at most two points, with neither consistently ahead.
\rev{Nor does the untied output head explain all of it: after one step, the same weights are level with or below the source baseline on seven of nine benchmarks (Table~\ref{tab:looped-dense-one-step}).}

Looped Qwen still trails Qwen3-4B-Base on every benchmark\rev{, so the conversion
improves} an existing checkpoint without repeating pretraining \rev{but does not
replace a larger one}.

\begingroup\revcolor

\subsection{The recipe at larger scale}
\label{sec:scaled}

To test whether our training recipe remains stable at larger scale, we train \scaled{} using the same architecture and stability settings as in Section~\ref{sec:looplm-training-recipe}. The model retains the shared 24-layer block, $T{=}4$, $\beta{=}0.15$, and a 1K-step learning-rate warmup. It uses the Qwen3 tokenizer, increasing the parameter count to 1.86B, and a modified data mixture that includes Russian text. Pretraining consists of 1.26T tokens at 4K context length, followed by 220B tokens at 16K and 32K, for a total of 1.48T tokens. The model then undergoes post-training (Appendix~\ref{app:scaled}).

The exit gate remains stable throughout training. After 1.26T tokens, the mean exit probabilities are $(0.15, 0.25, 0.30, 0.30)$ and change by less than 0.01 over the next 200B tokens at 16K context length. This extends the stability findings of Section~\ref{sec:looplm-stability} to a larger token budget, a different tokenizer, and a modified data mixture.

\begin{table}[!htbp]
\revcolor
\caption{Post-trained \scaled{} compared with published models, evaluated using the same harness (Appendix~\ref{app:scaled}). This comparison is not controlled. AIME uses pass@10, GPQA uses the Diamond subset, LCB denotes LiveCodeBench, and BFCL refers to its parallel category.}
\label{tab:scaled-results}
\begin{center}
\setlength{\tabcolsep}{3pt}
\resizebox{\linewidth}{!}{%
\begin{tabular}{lcccccccccccccc}
\toprule
 & Pretraining & MMLU & GPQA & GSM8K & MATH & AIME & AIME & Olympiad & Human & MBPP+ & LCB & IF & IF & BFCL \\
Model & tokens & Pro & & & 500 & 2025 & 2026 & Bench & Eval+ & & & Eval & Bench & \\
\midrule
\scaled{}          & 1.48T & 53.1 & 27.0 & 91.8 & 91.0 & 66.5 & 67.3 & 63.6 & 78.2 & 60.8 & 28.2 & 71.9 & 22.3 & 89.5 \\
Ouro-1.4B-Thinking & 7.7T  & 65.1 & 25.6 & 93.4 & 91.6 & 66.2 & 72.9 & 65.7 & 86.3 & 71.2 & 42.1 & 51.4 & 18.3 & 71.5 \\
Ouro-2.6B-Thinking & 7.7T  & 72.6 & 42.7 & 94.8 & 96.0 & 73.7 & 82.4 & 71.3 & 88.0 & 76.6 & 50.3 & 61.4 & 18.7 & 80.0 \\
Qwen3.5-2B         & n/a   & 65.6 & 53.0 & 84.5 & 74.4 & 39.3 & 43.9 & 43.0 & 62.0 & 50.7 & 9.1  & 79.7 & 37.7 & 83.0 \\
SmolLM3-3B         & 11.2T & 57.8 & 26.3 & 91.7 & 92.8 & 71.1 & 75.5 & 66.5 & 81.6 & 58.1 & 31.3 & 68.2 & 21.7 & 66.0 \\
\bottomrule
\end{tabular}}
\end{center}
\end{table}

Despite using 5.2 times fewer pretraining tokens than Ouro-1.4B-Thinking, \scaled{} achieves comparable scores on MATH-500, AIME 2025, and GPQA, and comes within approximately two points on GSM8K and OlympiadBench. It also performs better on instruction following and function calling, but trails Ouro on MMLU-Pro, AIME 2026, and code benchmarks.

Compared with dense models, \scaled{} remains competitive on mathematical reasoning, although it uses 5.24B parameters per token, compared with 1.88B and 3.08B for the dense models. These comparisons are not controlled because the models differ in training data, tokenizers, and post-training procedures.

We therefore also evaluate the larger run against dense baselines trained under matched conditions (Appendix~\ref{app:scaled}). At 419B tokens, LoopLM performs between its 24-layer and 72-layer dense counterparts on all 11 benchmarks, consistent with Table~\ref{tab:looplm-benches-anneal}, although the gap to the larger dense model is greater on mathematical reasoning.

\endgroup

\begingroup\revcolor
\subsection{Inference}
\label{sec:inference}

At three recurrent iterations, LoopLM applies 72 layers per token, matching the inference compute of \texttt{dense-ouro-3.9B}. Both models also require the same KV-cache capacity, since each recurrent iteration maintains its own keys and values. In vLLM benchmarks on a single H100 (Appendix~\ref{app:compute}), they achieve comparable decoding speeds across all batch sizes. However, LoopLM stores 4.6 GiB fewer model weights, leaving more memory for the KV cache and allowing 30 sequences of 4,096 tokens instead of 28.

Recurrence also enables two inference optimizations.

\textbf{Self-speculative decoding.}
An earlier iteration of LoopLM can generate draft tokens for the full model, eliminating the need to train a separate draft model. This is analogous to early-exit self-speculation in dense models \citep{elhoushi2024layerskip,zhang2024draftverify}. On the checkpoint from Table~\ref{tab:looplm-benches-anneal}, an average of 3.1 out of 4 draft tokens are accepted. At batch size 1, this accelerates four-iteration decoding by 1.65$\times$ compared with the same model using an FP8 cache without speculation, while preserving GSM8K accuracy. On the larger run, speculative decoding combined with an FP8 cache achieves a 1.7--2.2$\times$ speedup over standard bf16 decoding without speculation.

Our implementation uses a separate copy of the model weights and a one-iteration KV cache for drafting, so its memory overhead is comparable to that of a separate draft model of the same size. Against \texttt{dense-ouro-3.9B} using a separately trained \texttt{dense-ouro-1.4B} draft model, three-iteration LoopLM achieves similar decoding throughput (245 versus 251 tokens/s at batch size 1), with 4.6 GiB fewer model weights but 2.8 points lower GSM8K accuracy (Table~\ref{tab:spec}).

\textbf{KV-cache sharing.}
Recurrent iterations can also share parts of their KV cache. However, directly reusing the final iteration's cache, which \citet{zhu2025ouro} report as nearly lossless, causes degradation on long reasoning traces in our larger model. A more conservative strategy keeps the first iteration's cache separate while sharing the cache across the remaining three iterations. This reduces KV-cache memory by half, with a 3.7-point decrease on 300 few-shot GSM8K problems, within sampling noise (Appendix~\ref{app:scaled}).

\endgroup

\begingroup\revcolor
\section{Limitations}
\label{sec:limitations}

\textbf{Statistical uncertainty.}
Each configuration was trained with a single random seed. Consequently, the standard errors reported in Tables~\ref{tab:looplm-benches-anneal} and~\ref{tab:looped-dense-instruct} reflect test-set sampling uncertainty but do not account for variation across training runs.

\textbf{Training data and evaluation.}
The mid-training mixture includes training splits of several evaluated benchmarks. Although we decontaminated the mixture against their test sets and used identical data for all controlled comparisons, comparisons with published models (Table~\ref{tab:looplm-external}) do not share these controls.

\textbf{Training schedule.}
We did not optimize the allocation of tokens between pretraining and mid-training. Mid-training accounts for 32\% of our 310B-token budget, compared with approximately 1--6\% for OLMo~2 \citep{olmo2024olmo2} and 2\% for SmolLM3 \citep{bakouch2025smollm3}. The sensitivity of our results to this allocation remains unexplored.

\textbf{Language modeling performance.}
After instruction-formatted mid-training, LoopLM underperforms its parameter-matched dense baseline on LAMBADA and WikiText, despite outperforming it on raw-text checkpoints from the larger run (Appendix~\ref{app:lm}).

\textbf{Training compute.}
Our controlled comparisons match training data and token budgets, but not training compute. Under equal training compute, the parameter-matched dense baseline could process approximately four times as many tokens, while the inference-compute-matched baseline could process 1.4 times as many. We did not evaluate these additional training budgets.

\endgroup

\section{Conclusion}

This work gives two practical ways to build a looped language model: a short recipe for training one from scratch, and a conversion of a pretrained dense model with minimal architectural changes.

For from-scratch training, a two-stage recipe trains a looped model at a fraction of the token budget of earlier multi-stage pipelines, \rev{and a learning-rate warmup prevents early gate collapse at a larger KL coefficient, which alone} lets the gate recover if it does collapse.
The resulting LoopLM improves over a parameter-matched dense model on \rev{all 12 benchmarks of Table~\ref{tab:looplm-benches-anneal}} and, at equal inference compute, comes within about a point of a dense model with \rev{2.7} times its parameters on math and reading comprehension; the larger model \rev{leads on knowledge and commonsense benchmarks}.

For pretrained models, applying \rev{the Qwen3-1.7B-Base decoder} repeatedly does not produce refinement.
One learned scalar that mixes the token embedding into each recurrent step, together with a smoothed exit loss, is enough to turn Qwen3-1.7B-Base into Looped Qwen, which improves on the source checkpoint given the same continued training \rev{on most benchmarks}, with no step-specific parameters.

Both results are measured against dense baselines trained on the same data \rev{and token budget. At fixed data, recurrence buys parameter efficiency and does not buy training-compute efficiency: the 3.9B twin with the same inference compute stays ahead or level and costs less to train. A larger run with the same stability settings keeps the exit gate stable (Section~\ref{sec:scaled}).}


\section*{AI use statement}

Generative AI tools were used to assist with drafting and polishing the
manuscript, retrieval and discovery of relevant literature, and refactoring
author-written code. In particular, AI assistance was used to improve wording,
clarity, and organization of the text, to help identify potentially relevant
prior work, and to restructure, simplify, and improve the readability of
existing implementation code without changing the intended algorithms or
experimental methodology. All retrieved references were independently checked
by the authors against the original sources, and all AI-assisted code changes
were reviewed and tested by the authors before use.

Generative AI was not used to generate synthetic data, formulate the scientific
hypotheses or methodology, derive theoretical claims or proofs, conduct data
cleaning or analysis, interpret experimental results, or translate the
manuscript. It was not used to generate experimental results or make scientific
conclusions on behalf of the authors.

All AI-assisted text and code were reviewed, edited, and verified by the
authors. Numerical results, experimental claims, citations, and descriptions of
prior work were checked against the corresponding experiments, source
materials, and original publications. The authors take full responsibility for
the accuracy, integrity, and final content of the paper.

\section*{Ethics statement}

We adhere to the ICLR Code of Ethics. This work studies the architecture,
training, and evaluation of language models and does not involve human-subject
experiments or the collection of new personal or sensitive data. Training uses
existing language-model pretraining and instruction-tuning datasets; we do not
release additional user data or construct datasets containing newly collected
personal information.

The models studied in this work are general-purpose language models and
therefore inherit the usual potential risks associated with such systems,
including generation of inaccurate, biased, or harmful content and possible
dual-use applications. Our work does not introduce mechanisms specifically
designed for harmful use, nor does it evaluate or optimize models for such
applications. The proposed methods primarily change how computation is reused
across recurrent steps and do not remove the need for the same safety,
fairness, privacy, and deployment considerations that apply to conventional
language models.

To the best of our knowledge, this work raises no additional legal, privacy, or
fairness concerns beyond those associated with the underlying public
pretraining and instruction-tuning corpora and pretrained checkpoints. Any
funding sources and potential conflicts of interest will be disclosed in the
camera-ready version as required by the conference.

\section*{Reproducibility statement}
The from-scratch LoopLM is specified in Table~\ref{tab:our-recipe} and Appendix~\ref{app:looplm-setups}: architecture, recurrent objective, optimizer, schedules and both data mixtures. The Looped Qwen conversion is specified in Section~\ref{sec:looped-dense} and Appendix~\ref{app:looped-dense-setups}: architecture, recurrent modules, optimizer, seeds and the curated data mixture; the web-scale run is reported in Appendix~\ref{app:looped-dense-web}.

For every controlled comparison, the looped model and its dense baselines draw on the same data sources and use the same training stages, token budget and evaluation procedure (Section~\ref{sec:baselines}). The depth study (Section~\ref{sec:looplm-stability}) and the shared-versus-LoRA comparison (Section~\ref{sec:looped-dense-results}) are in the main text; the exit-gate stability sweep, the normalization ablation and the reinforcement-learning runs are in Appendices~\ref{app:early-collapse}, \ref{app:norm-ablation} and~\ref{app:rl}. \rev{Appendix~\ref{app:tah} reports our attempt to reproduce Think-at-Hard.}

The supplementary material contains anonymized training code for both tracks, with one-GPU smoke tests; the from-scratch trainer also trains the dense twins. \rev{It also contains configuration files for every training stage in the paper, including the larger run of Section~\ref{sec:scaled}. Table~\ref{tab:eval-protocol} lists the evaluation settings of Table~\ref{tab:looplm-benches-anneal}, and Appendix~\ref{app:looped-dense-setups} gives those of Table~\ref{tab:looped-dense-instruct}. The dense Qwen baselines and the larger run were trained with trainers that are not included, and their configuration files record the settings.} The LoopLM smoke test runs on synthetic data, and the Looped Qwen test also needs the public Qwen3-1.7B-Base weights. \rev{The supplementary material does not include the training data, data preparation, the evaluation code, the normalization variants of Appendix~\ref{app:norm-ablation} or the reinforcement-learning code.}

\bibliography{iclr2027_conference}
\bibliographystyle{iclr2027_conference}
\clearpage
\appendix

\section{Layer-normalization ablation}
\label{app:norm-ablation}

Weight sharing makes normalization unusually important in recurrent models:
the same decoder block is traversed up to $T$ times, and its parameters receive
gradients accumulated across recurrent iterations. Prior work has likewise
identified normalization as an important stability choice in deeply recurrent
Transformers: HRM-Text uses a parameter-free pre-norm variant, MagicNorm
\citep{wang2026hrmtext}, while recurrent-depth pretraining has shown strong
sensitivity to normalization placement \citep{geiping2025scaling}.

We ablate two choices separately:
(i) normalization inside the recurrent decoder block
(\emph{decoder norm}), and
(ii) normalization of the hidden state before the shared LM head
(\emph{final norm}).

Unless stated otherwise, all runs use the architecture and optimizer of our
main recipe, $T{=}4$, a 500K-token batch, and a matched budget of 19K steps.

\subsubsection*{Normalization schemes}

For the decoder block, we compare five variants.
\emph{Ouro-norm} is the configuration used by \citet{zhu2025ouro} and by our
main recipe (Section~\ref{sec:looplm-training-recipe}): it normalizes the input
and the output of each sub-layer and adds the normalized output to the residual stream.
\emph{Pre-norm} applies normalization before each attention or MLP sub-layer,
followed by the residual addition.
\emph{Post-norm} normalizes after the residual addition.
\emph{Sandwich-norm} also normalizes the input of each sub-layer, but applies the
second normalization after the residual addition, so the residual stream itself
is normalized, following \citet{geiping2025scaling}.
Finally, \emph{Magic-norm} is pre-norm with the learnable normalization scale
removed, following \citet{wang2026hrmtext}.

For the final hidden state, we test learnable RMSNorm (\emph{RMS}),
parameter-free RMSNorm (\emph{FRMS}), and no normalization
(\emph{identity}).

\begin{table}[H]
\centering
\caption{LM loss after 19K training steps with a 500K-token batch.}
\label{tab:norm-ablation-main}
\small
\setlength{\tabcolsep}{7pt}
\renewcommand{\arraystretch}{1.05}
\begin{tabular}{lccc}
\toprule
& \multicolumn{3}{c}{\textbf{Final normalization}} \\
\cmidrule(lr){2-4}
\textbf{Decoder norm} &
\textbf{RMS} &
\textbf{FRMS} &
\textbf{Identity} \\
\midrule
Ouro-norm (baseline) & 1.98 & --   & 2.11 \\
Pre-norm             & 2.03 & 2.07 & 2.11 \\
Magic-norm           & --   & 2.04 & 2.07 \\
\midrule
Post-norm            & --   & --   & 7.37 \\
Sandwich-norm        & --   & --   & 7.40 \\
\bottomrule
\end{tabular}
\end{table}

The result separates cleanly into two groups \rev{(Table~\ref{tab:norm-ablation-main})}. Ouro-norm, pre-norm, and
MagicNorm all train normally and finish in a narrow $1.98$--$2.11$ range.
In contrast, post-norm and sandwich-norm plateau near $7.4$ and fail to train.

\begin{figure}[t]
\centering
\includegraphics[width=0.72\textwidth]{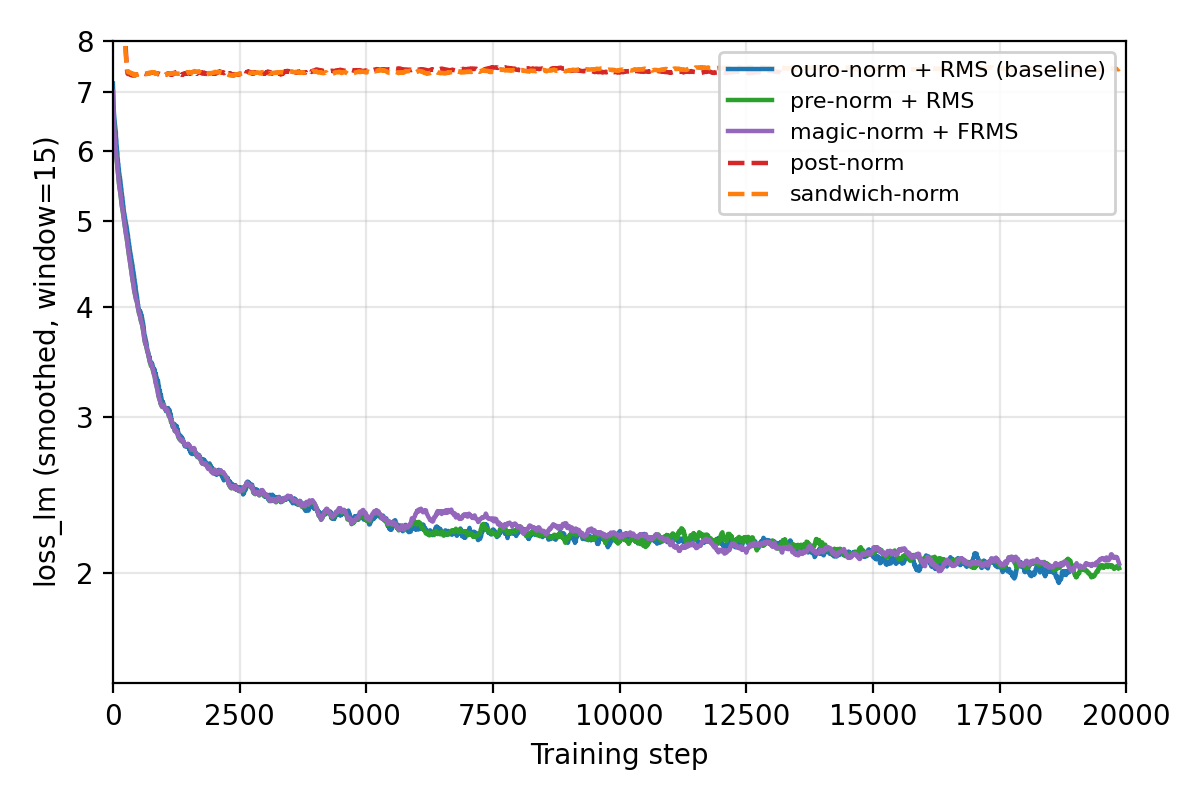}
\caption{
Training LM loss (15-step moving average) for representative pre-norm-family
variants and for post-norm and sandwich-norm. The pre-norm-family curves nearly
overlap, while post-norm and sandwich-norm plateau early.
}
\label{fig:norm-ablation-loss}
\end{figure}

Figure~\ref{fig:norm-ablation-loss} shows that the divergence appears early.
Representative pre-norm-family runs converge along nearly identical
trajectories, whereas post-norm and sandwich-norm plateau within roughly the
first 1K steps and never recover.

\subsubsection*{Why post-norm fails}

\begin{figure}[t]
\centering
\includegraphics[width=0.72\textwidth]{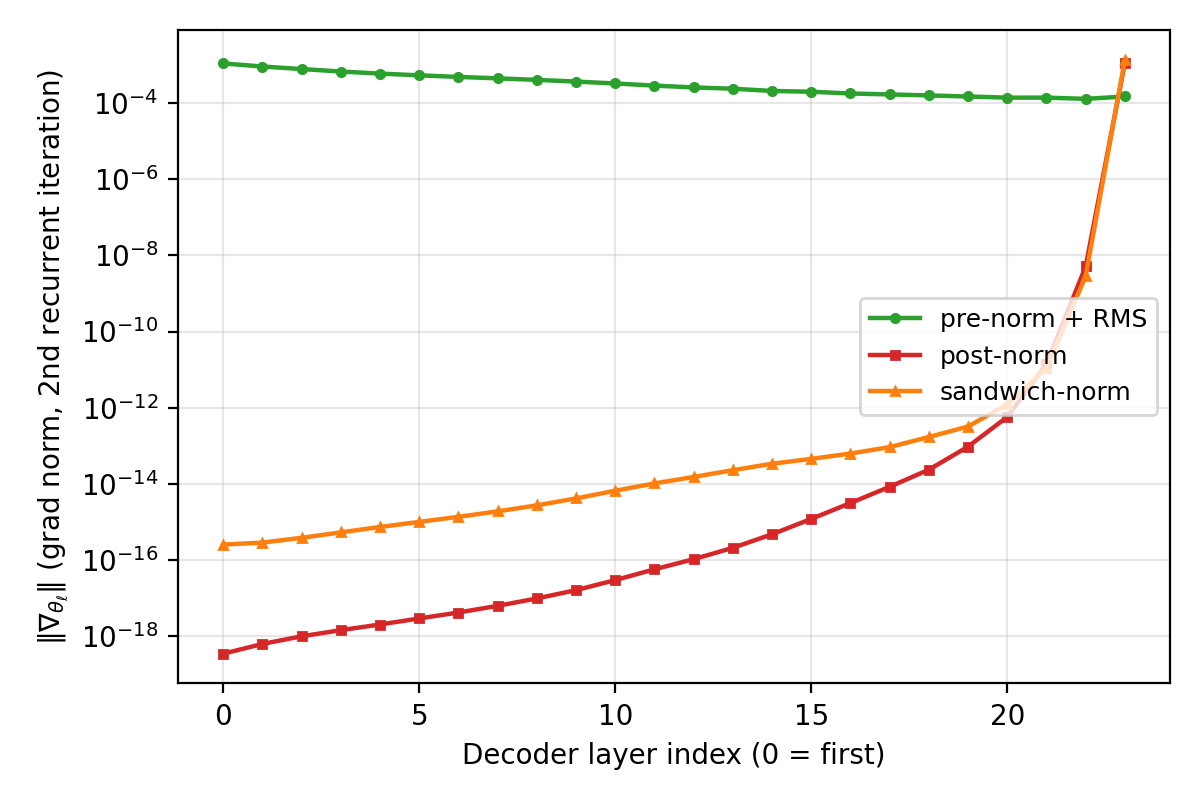}
\caption{
Gradient norm by decoder-layer depth at \rev{the second} recurrent iteration.
Pre-norm preserves gradients throughout the block, whereas post-norm and
sandwich-norm concentrate almost all gradient in the final decoder layer.
}
\label{fig:norm-ablation-grad}
\end{figure}

To localize the failure, we log the parameter-gradient norm of every decoder
layer separately for each recurrent iteration.
Figure~\ref{fig:norm-ablation-grad} shows the depth profile at the second
iteration.

With pre-norm, the gradient decreases by only about $8\times$ from the first
decoder layer to the last. With post-norm and sandwich-norm, however, layers
0--22 receive gradients that are \rev{$5$}--$16$ orders of magnitude smaller than
the gradient of layer 23. Almost all learning is therefore concentrated in the
final layer of the shared block. In effect, a nominally 24-layer recurrent
block behaves during optimization much closer to a one-layer block.

\subsubsection*{Post-norm mitigation}

We test whether the post-norm failure can be alleviated by changing the
optimization schedule or shortening the recurrent path.

Increasing learning-rate warmup from 1K to 10K steps does not help: post-norm
still converges to the same loss plateau near $7$.

\begin{figure}[t]
\centering
\includegraphics[width=0.72\textwidth]{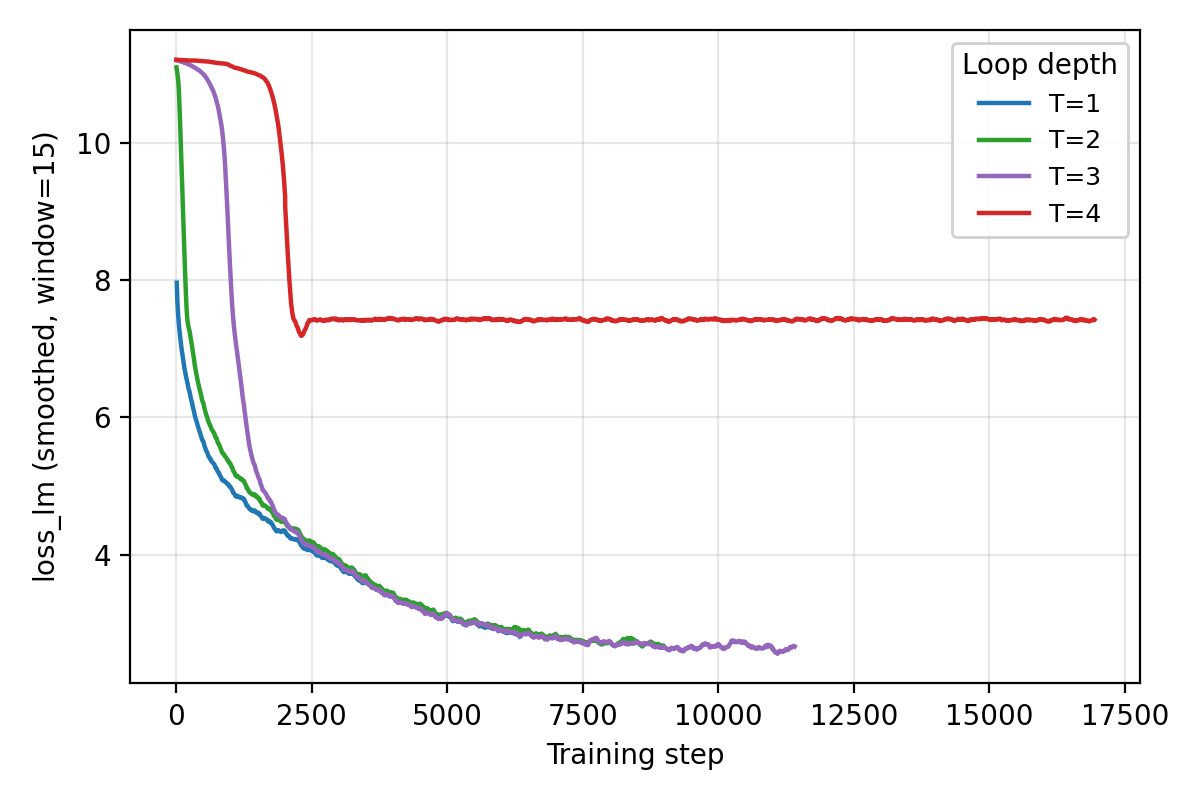}
\caption{
Post-norm training LM loss (15-step moving average) as a function of recurrent
depth $T$, using a 500K-token batch and a 10K-step warmup. Lower recurrent
depth mitigates the failure, but remains worse than the pre-norm-family
configurations.
}
\label{fig:norm-ablation-post-depth}
\end{figure}

Reducing recurrent depth provides a partial workaround.
Figure~\ref{fig:norm-ablation-post-depth} compares post-norm at
$T \in \{1,2,3,4\}$ with a 10K-step warmup and otherwise identical settings.
The $T{=}1$--$3$ runs converge to losses between $2.66$ and $2.82$.
The $T{=}4$ run \rev{stays near its initial loss} for roughly 2K steps,
\rev{then falls only to} the $\sim7.4$ plateau.

Thus, shortening the recurrent path avoids the most severe optimization
failure, but does not recover the performance of the pre-norm family.

\subsubsection*{Final normalization}

Once the decoder uses a stable pre-norm-family configuration, the exact choice
of final normalization has only a small effect on mean loss, but a much larger
effect on training stability.

Removing the final norm increases the matched-budget loss by \rev{$0.03$}--$0.13$
and introduces repeated loss spikes. We define a spike as a step-over-step
loss increase greater than $0.5$ above a 15-step rolling median, and count
spikes over the first 25K training steps.

\begin{table}[H]
\centering
\caption{
Effect of final normalization on loss and training stability.
Loss is measured at 19K steps; spikes and maximum gradient norm are measured
over the first 25K steps.
}
\label{tab:norm-ablation-final-norm}
\small
\setlength{\tabcolsep}{6pt}
\renewcommand{\arraystretch}{1.05}
\begin{tabular}{llccc}
\toprule
\textbf{Decoder norm} &
\textbf{Final norm} &
\textbf{Loss} &
\textbf{\#Spikes} &
\textbf{Max grad norm} \\
\midrule
Ouro-norm  & RMS      & 1.98 & 0  & $42.3$ \\
Ouro-norm  & Identity & 2.11 & 5  & $1.7\times10^{3}$ \\
Pre-norm   & RMS      & 2.03 & 0  & $15.7$ \\
Pre-norm   & FRMS     & 2.07 & 0  & $28.4$ \\
Pre-norm   & Identity & 2.11 & 11 & $5.2\times10^{7}$ \\
Magic-norm & FRMS     & 2.04 & 0  & $30.0$ \\
Magic-norm & Identity & 2.07 & 10 & $1.3\times10^{6}$ \\
\bottomrule
\end{tabular}
\end{table}

All RMS/FRMS variants have zero spikes, whereas the identity variants produce
5--11 \rev{(Table~\ref{tab:norm-ablation-final-norm})}. The difference is even clearer in gradient magnitude: with a final norm,
the peak total gradient norm stays between $15.7$ and $42.3$; without one, it
reaches as high as $5.2\times10^{7}$.

Whether the final RMSNorm has a learnable scale matters much less.
For pre-norm, replacing RMS with FRMS changes the loss from $2.03$ to $2.07$,
and both runs remain stable. The important choice is therefore whether a final
normalization is present, not whether its scale is learned.

\subsubsection*{Robustness checks}

Finally, we repeat the main normalization comparison at the 2M-token batch size
used in our full training recipe. The conclusion does not change:
post-norm, sandwich-norm, and a parameter-free post-norm variant all plateau at
$7.40$--$7.41$, closely matching the 500K-token experiments.

Removing the learnable scale from post-norm therefore does not restore gradient
flow. Together with the warmup and recurrent-depth experiments, this indicates
that the failure is primarily caused by normalization placement rather than by
batch size, warmup length, or the presence of learnable normalization
parameters.

\subsubsection*{Takeaway}

Normalization placement around the shared recurrent block matters much more
than the exact normalization formula. Post-norm and sandwich-norm fail because
gradients vanish through almost the entire decoder block, and this failure
persists across warmup lengths, batch sizes, and normalization
parameterizations.

Within the stable pre-norm family, ouro-norm, standard pre-norm, and MagicNorm
perform similarly. A final normalization before the LM head is important for
stability, while learnable and parameter-free RMSNorm behave similarly.

Pre-norm+RMS is therefore a simple alternative to the ouro-norm configuration
used in our main experiments, reaching a very similar final loss
($2.03$ vs.\ $1.98$). We retain ouro-norm in the main recipe because the
large-scale experiments had already been completed before this ablation.

\section{Early pretraining collapse to a dense regime}
\label{app:early-collapse}

Section~\ref{sec:looplm-stability} shows that learning-rate warmup \rev{prevents} the
early gate collapse \rev{at our $\beta$}. Here we examine the same failure mode more closely in
the regime where it \rev{is common}: training without warmup. Rather than
repeating the warmup $\times$ $\beta$ comparison from
Figure~\ref{fig:looplm_stability_analysis}, we sweep $\beta$ to characterize
when an early collapse becomes permanent and when the model recovers.

All runs use the same 2M-token-batch configuration as the main recipe
(Table~\ref{tab:our-recipe}). We train without warmup at
$\beta \in \{0.05, 0.10, 0.50\}$ using a constant learning rate, and include
one no-warmup cosine-schedule run at $\beta{=}0.15$, whose cosine spans a long budget, so its learning rate stays near the peak over the steps shown. We track the probability
of exiting after the first recurrent step, $p_0$, which directly measures how
close the model is to single-step dense training.

\begin{table}
\centering
\caption{
No-warmup $\beta$ sweep. Peak $p_0$ is the maximum first-step exit probability
after the first 50 steps; late $p_0$ is its mean over steps 5K to 6K. Loss is the
LM loss at step 2K, the latest step all four runs reached. The $\beta{=}0.05$ run
is the collapsed run shown in Figure~\ref{fig:looplm_stability_analysis}.
}
\label{tab:early-collapse-runs}
\small
\setlength{\tabcolsep}{5pt}
\renewcommand{\arraystretch}{1.05}
\begin{tabular}{l l r c c c l}
\toprule
$\boldsymbol{\beta}$ &
\textbf{Schedule} &
\textbf{Steps} &
\textbf{Peak $p_0$} &
\textbf{Late $p_0$} &
\textbf{Loss at 2K} &
\textbf{Outcome} \\
\midrule
0.05 & Constant & 2,267  & 1.00 & 1.00$^{*}$ & 2.38 & No recovery in 2.3K steps \\
0.10 & Constant & 54,558 & 1.00 & 0.11 & 2.36 & Recovers \\
0.15 & Cosine   & 24,579 & 0.99 & 0.14 & 2.39 & Recovers \\
0.50 & Constant & 27,204 & 0.66 & 0.20 & 2.59 & Never collapses \\
\bottomrule
\multicolumn{7}{l}{\footnotesize $^{*}$Mean over the whole run, which was stopped at step 2,267.} \\
\end{tabular}
\end{table}

\begin{figure}
\centering
\includegraphics[width=0.85\linewidth]{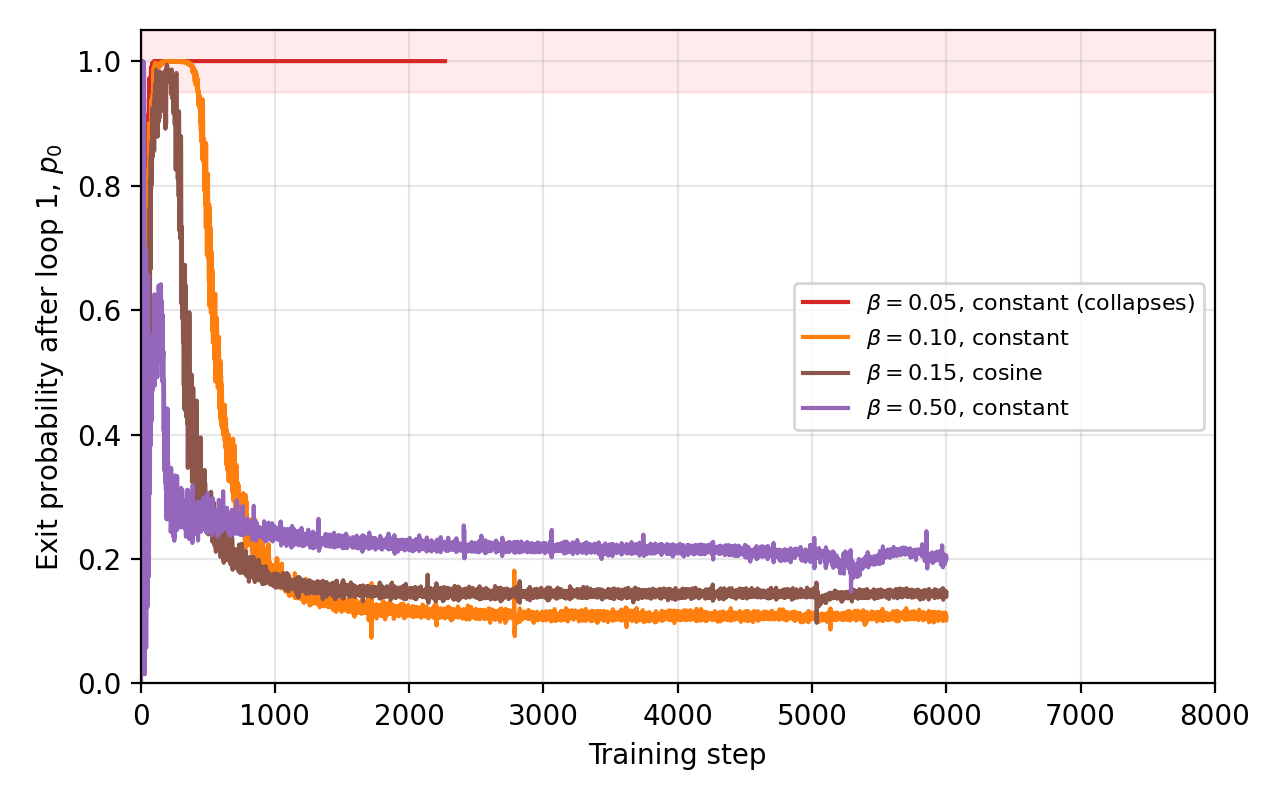}
\caption{
First-step exit probability $p_0$ during the first \rev{6K} training steps without
warmup. The shaded region marks $p_0>0.95$. The $\beta{=}0.05$ run stays
collapsed until it was stopped at step 2,267; $\beta{=}0.10$ and $0.15$ recover,
and $\beta{=}0.50$ never enters the shaded region after its first 50 steps.
}
\label{fig:early-collapse-gate}
\end{figure}

\subsubsection*{Collapse is usually reversible}

The sweep \rev{(Table~\ref{tab:early-collapse-runs} and Figure~\ref{fig:early-collapse-gate})} shows that entering the dense regime does not by itself imply a
training failure. Three of the four runs enter the collapsed regime early in
training: at $\beta \in \{0.05, 0.10, 0.15\}$ the first-step exit probability
reaches 0.99 to 1.00 within the first few hundred steps. At $\beta{=}0.10$ and
$\beta{=}0.15$ the gate then returns to a recurrent regime, with late $p_0$ of
0.11 and 0.14; both had recovered by about step 1K. At $\beta{=}0.05$ it stays at
$p_0{=}1.00$ for all 2,267 steps of the run. At $\beta{=}0.50$ the gate never
enters the collapsed regime: after the first 50 steps $p_0$ peaks at 0.66.
Without warmup, a collapse is therefore the usual start of training at
moderate $\beta$, and $\beta$ decides whether the gate recovers from it. At
step 2K the collapsed run has the same LM loss as the recovered ones (2.38
against 2.36 and 2.39), so the loss alone does not reveal the collapse.

\subsubsection*{The recovered exit distribution depends on $\beta$}

The runs that train in the recurrent regime do not converge to the same exit
distribution. Their late first-step exit probabilities are $0.11$, $0.14$, and
$0.20$ for $\beta \in \{0.10,0.15,0.50\}$, respectively.

This trend is consistent with the KL regularizer pulling the exit distribution
toward the uniform $T{=}4$ prior, for which $p_0=0.25$. Larger $\beta$
therefore both helps the gate recover from, or avoid, the early first-step
collapse and places more probability on earlier exits after training has
stabilized.

The role of warmup is different: at $\beta{=}0.15$ it keeps the gate out of the
collapsed regime ($p_0>0.95$) from the start, and at $\beta{=}0.05$ it limits that
regime to the first few dozen steps. Without warmup, $\beta$ primarily
determines whether the gate can recover once the collapse has occurred.

\subsubsection*{Gate collapse is distinct from normalization failure}

This failure mode is different from the normalization failure analyzed in
Appendix~\ref{app:norm-ablation}. Here, the shared decoder block continues to
receive gradients and train; what disappears is the contribution of later
recurrent steps because the exit gate places nearly all probability on the
first step.

Under post-norm and sandwich-norm, by contrast, the gate is not the primary
problem. Gradients vanish through almost the entire decoder block, leaving
essentially only its final layer trainable. That failure is persistent and is
not repaired by changing $\beta$ or extending warmup.

The two cases therefore correspond to different forms of effective model
collapse: gate collapse reduces a recurrent model to single-step computation,
temporarily or permanently, whereas normalization failure reduces a deep
recurrent block to effectively single-layer optimization.

\section{When does RL help a looped model?}
\label{app:rl}

We ran RLTT on our 1.4B LoopLM in six configurations, varying the starting
checkpoint, target task, and learning rate. All runs used temperature $0.9$,
groups of 8 rollouts, 32 prompts per optimizer step, and a fixed recurrent depth
of $T{=}4$. Two runs used a learning rate of $10^{-5}$ and were evaluated with
full benchmark sweeps; the remaining four used the RLTT reference learning rate
of $10^{-6}$.

\subsubsection*{\rev{RL did not help once the rollout task was mostly solved}}

Only one of the six runs produced a clear downstream improvement. Starting from
the pretraining checkpoint and optimizing GSM8K, RLTT improved 5-shot GSM8K
from $35.3$ to $47.3$ in 140 steps, a gain of $12.0$ points \rev{(Table~\ref{tab:rl-results})}.

In contrast, optimizing MATH from a mid-trained checkpoint produced no measurable
gain: MATH changed from $55.7$ to $55.2$, while the GSM8K guard metric changed
from $73.5$ to $73.2$, both within evaluation error. \rev{This checkpoint is step 33,000 of the mid-training of an earlier run of this recipe, not
the checkpoint of Table~\ref{tab:looplm-benches-anneal}, so its scores are lower.}

\rev{A key} difference between these runs is the solve rate of the policy during
rollout. Over the first 20 optimizer steps, the pretraining checkpoint solved
about $30\%$ of its GSM8K rollouts. Most groups therefore contained both correct
and incorrect samples, producing a useful group-relative learning signal.
The mid-trained checkpoint already solved about $77\%$ of its MATH rollouts.
Many groups were consequently uniform or nearly uniform, causing the
group-relative advantage to vanish.

\begin{table}
\centering
\caption{
RLTT results at learning rate $10^{-5}$ after 140 optimizer steps.
The reported uncertainty is the \texttt{lm-evaluation-harness} standard error
and depends on the benchmark rather than the checkpoint.
GSM8K in the MATH run is a guard metric, not an RL target.
\rev{The pretraining checkpoint is evaluated with the standard 5-shot GSM8K task. The MATH run
starts from an earlier mid-training run than Table~\ref{tab:looplm-benches-anneal}, which explains
its lower scores.}
}
\label{tab:rl-results}
\small
\setlength{\tabcolsep}{5pt}
\renewcommand{\arraystretch}{1.05}
\begin{tabular}{lccc}
\toprule
\textbf{Benchmark} &
\textbf{Base} &
\textbf{Step 100} &
\textbf{Step 140 ($\Delta$)} \\
\midrule
\multicolumn{4}{l}{
\emph{GSM8K target; pretraining checkpoint; solve rate $\approx0.30$}
} \\
GSM8K, 5-shot (target)
& $35.33 \pm 1.32$
& $46.78$
& $47.31$ ($+11.98$) \\
MMLU, 5-shot (guard)
& $47.59 \pm 0.41$
& $48.03$
& $47.99$ ($+0.40$) \\
HellaSwag, 10-shot (guard)
& $67.59 \pm 0.47$
& $68.08$
& $68.09$ ($+0.50$) \\
\midrule
\multicolumn{4}{l}{
\emph{MATH target; mid-trained checkpoint; solve rate $\approx0.77$}
} \\
MATH, 0-shot (target)
& $55.68 \pm 0.68$
& --
& $55.18$ ($-0.50$) \\
GSM8K, 0-shot (guard)
& $73.46 \pm 1.22$
& --
& $73.16$ ($-0.30$) \\
\bottomrule
\end{tabular}
\end{table}

\subsubsection*{Training telemetry predicts whether RL is moving the policy}

The two runs separate clearly before any downstream evaluation.
Table~\ref{tab:rl-telemetry} reports KL divergence from the reference policy
and total gradient norm over the first and last 20 optimizer steps.

In the successful GSM8K run, KL grows from $0.013$ to $0.084$, while the
gradient norm remains close to $0.6$. In the MATH run, KL reaches only $0.003$
and the gradient norm stays near $0.1$. The policy therefore remains close to
its initialization throughout training.

These quantities provide a cheap early diagnostic. A growing KL together with a
non-negligible gradient norm indicates that RLTT is actually moving the policy;
persistently tiny values suggest that a run is unlikely to produce a downstream
gain.

\begin{table}
\centering
\caption{
Training telemetry for the two $10^{-5}$ runs, averaged over the first and last
20 optimizer steps. KL is measured against the reference policy and therefore
starts at zero by construction.
}
\label{tab:rl-telemetry}
\small
\setlength{\tabcolsep}{7pt}
\renewcommand{\arraystretch}{1.05}
\begin{tabular}{lcc}
\toprule
&
\textbf{GSM8K run} &
\textbf{MATH run} \\
\midrule
KL, first 20 steps
& $0.013$ & $<0.001$ \\
KL, last 20 steps
& $0.084$ & $0.003$ \\
Gradient norm, first 20 steps
& $0.58$ & $0.12$ \\
Gradient norm, last 20 steps
& $0.64$ & $0.11$ \\
\bottomrule
\end{tabular}
\end{table}

\subsubsection*{The policy must move before task difficulty matters}

Learning rate is a second necessary condition. At $10^{-6}$, none of the four
additional runs meaningfully departed from the reference policy, regardless of
the starting checkpoint or target task. After 94--421 optimizer steps, KL
remained near $2\times10^{-4}$.

One of these runs started from the pretraining checkpoint and optimized MATH,
providing a case in which the target task was substantially harder for the
initial policy. However, because the smaller learning rate failed to move the
policy, this experiment does not test whether RLTT is effective when the rollout
solve rate is very low. The lower end of the useful solve-rate range therefore
remains unresolved.

\subsubsection*{The useful update is short and does not hurt guard metrics}

The GSM8K improvement also saturates quickly. Accuracy reaches $46.8$ at step
100 and $47.3$ at step 140, a difference smaller than the benchmark standard
error. Continuing the same run therefore provides little additional benefit.

Neither guard metric \rev{falls}: MMLU moves by $+0.4$ points
and HellaSwag by $+0.5$. Within this experiment, the GSM8K improvement therefore
does not come with measurable forgetting on these benchmarks.

Taken together, these runs suggest two practical requirements for RLTT in this
setting. First, the learning rate must be large enough for the policy to leave
its initialization. Second, the rollout task should produce mixed-success
groups so that group-relative rewards contain useful variation. Our successful
run satisfies both conditions; the high-solve-rate MATH run fails the second,
while all $10^{-6}$ runs fail the first.

Whether RLTT can improve an already mid-trained checkpoint on a task selected
specifically to produce a mixed rollout solve rate remains an open question.

\begingroup\revcolor
\subsubsection*{The exit gate can also collapse under RL}
On the larger run of Section~\ref{sec:scaled}, we ran GRPO with the RLTT loss on MATH from the 16K-context base checkpoint (step 35,000 of that stage), with a 4-shot prompt, 8 rollouts per prompt and 140 optimizer steps.
In the first run the exit probabilities that weight the per-iteration losses carried gradient.
The mean probability of exiting at the last iteration fell from 0.29 at the start to 0.21 by step 60, 0.05 by step 95 and about 0.03 from step 106, so the model moved almost all of the last iteration's exit mass to the earlier iterations (Table~\ref{tab:rl-gate}).
Response length had already grown from about 500 to 3,000 tokens while the reward rose; after the collapse it reached 6,500 tokens at step 120, with three quarters of the responses truncated, and the mean training reward fell from 0.69 over steps 61--80 to 0.23 over steps 121--140.
The direction of the collapse is consistent with the loss.
The log-likelihood of the sampled tokens rose with the iteration on every one of the 140 steps, so for responses with negative advantage, the gradient through the exit probabilities moves weight to the earlier iterations.

A second run detached the exit probabilities from the gradient, and its last-iteration probability stayed between 0.28 and 0.31 for all 140 steps.
On MATH-500, with greedy decoding and the same prompt, the base checkpoint scores 25.8, the first run 41.8 at step 140, and the second run 68.0, 70.2 and 74.6 at steps 60, 100 and 140; on GSM8K the second run reaches 79.4 against 36.4 for the base checkpoint.
The second run also changed the loss aggregation, the KL coefficient, the gradient clip, the batch, the learning-rate schedule and the response limit, so these numbers do not isolate the effect of detaching.
They do show the collapse in the run where the exit distribution received the policy gradient, and the last iteration kept in use in the run where it did not.
This is the third setting, after pretraining (Section~\ref{sec:looplm-stability}) and conversion (Section~\ref{sec:looped-dense-stability}), in which the exit gate abandons some iterations unless something protects them.

\begin{table}[h]
\revcolor
\centering
\caption{RL on the larger run with and without gradient through the exit probabilities: probability of exiting at the last iteration ($p_3$), mean response length in tokens and share of truncated responses. The response limit is 8,192 tokens in the first run and 4,096 in the second.}
\label{tab:rl-gate}
\small
\setlength{\tabcolsep}{5pt}
\begin{tabular}{rccccccc}
\toprule
 & \multicolumn{3}{c}{Gradient through the exit probabilities} & & \multicolumn{3}{c}{Exit probabilities detached} \\
\cmidrule(lr){2-4}\cmidrule(lr){6-8}
Step & $p_3$ & Length & Truncated & & $p_3$ & Length & Truncated \\
\midrule
1 & 0.29 & 521 & 0.8\% & & 0.30 & 329 & 1.2\% \\
40 & 0.22 & 2{,}702 & 16.8\% & & 0.28 & 997 & 13.1\% \\
60 & 0.21 & 2{,}979 & 14.8\% & & 0.29 & 1{,}141 & 5.1\% \\
80 & 0.11 & 3{,}459 & 21.5\% & & 0.29 & 1{,}018 & 4.3\% \\
95 & 0.05 & 3{,}472 & 22.3\% & & 0.29 & 1{,}558 & 19.5\% \\
120 & 0.03 & 6{,}528 & 75.0\% & & 0.29 & 1{,}136 & 9.6\% \\
140 & 0.03 & 5{,}411 & 59.4\% & & 0.29 & 1{,}979 & 26.8\% \\
\bottomrule
\end{tabular}
\end{table}
\endgroup

\section{LoopLM architecture and training setup}
\label{app:looplm-setups}

Table~\ref{tab:our-recipe} summarizes the optimizer and schedule of our
two-stage recipe. This appendix provides the remaining details needed to
reproduce the from-scratch LoopLM experiments: the backbone, recurrent
objective, training mixtures, and dense baselines. We do not repeat the
motivation for choosing $T{=}4$ or the warmup analysis from
Sections~\ref{sec:looplm-training-recipe} and~\ref{sec:looplm-stability}.

\subsection{Shared-block architecture}

Our from-scratch model follows the 1.4B Ouro LoopLM architecture of
\citet{zhu2025ouro} \rev{(Table~\ref{tab:looplm-backbone})}. A 24-layer decoder block is reused for $T{=}4$ recurrent
steps, with a shared LM head and a learned per-token exit gate. Token embeddings
are introduced only once: later recurrent steps consume the hidden state from
the previous step, without embedding re-injection
(Appendix~\ref{app:embedding-injections}).

The main model uses the Ouro normalization layout. Each attention and MLP
sub-layer normalizes its input and its output, and the normalized output is
added to the residual stream, which is itself left unnormalized. A learnable RMSNorm is then applied to the hidden state after
each recurrent step, before both the exit gate and the LM head.
Appendix~\ref{app:norm-ablation} studies alternative normalization schemes.

\begin{table}
\centering
\caption{Backbone of the from-scratch 1.4B LoopLM, matching Ouro-1.4B
\citep{zhu2025ouro}.}
\label{tab:looplm-backbone}
\footnotesize
\setlength{\tabcolsep}{4pt}
\renewcommand{\arraystretch}{1.05}
\begin{tabular}{@{}p{0.34\linewidth}p{0.59\linewidth}@{}}
\toprule
\textbf{Parameter} & \textbf{Value} \\
\midrule
Initialization & Random, std.\ $0.02$ \\
Decoder layers & $24$, shared across recurrent steps \\
Hidden / FFN dimension & $2048$ / $5632$ \\
Attention heads & $16$ query, $16$ key--value; no GQA \\
Attention head dimension & $128$ \\
Attention pattern & Full causal attention; no sliding window \\
Activation & SiLU-gated feed-forward network \\
Decoder / final norm & Ouro-norm / learnable RMSNorm, $\epsilon=10^{-6}$ \\
Vocabulary size & $49{,}152$ (Ouro tokenizer) \\
Maximum positions / RoPE & $4096$ / base $4\times10^{4}$ \\
Input/output embeddings & Untied; LM head shared across steps \\
Trainable parameters & $1.4$B \\
\bottomrule
\end{tabular}
\end{table}

\subsection{Recurrence and exit objective}

Let $e$ denote the token embeddings and $B$ the shared decoder block. The model
applies the same block recurrently:
\begin{equation}
h_0 = e,
\qquad
h_t = B(h_{t-1}),
\quad t=1,\ldots,T.
\end{equation}

After each recurrent step, a shared linear gate produces a halt probability
\begin{equation}
\lambda_t = \sigma(w_g^\top h_t).
\end{equation}

Following \citet{zhu2025ouro}, these halt probabilities define the exit
distribution through a survival construction:
\begin{equation}
p_t =
\lambda_t \prod_{s<t}(1-\lambda_s)
\quad (t<T),
\qquad
p_T =
\prod_{s<T}(1-\lambda_s),
\end{equation}
with the final distribution normalized to sum to one.
\rev{Elsewhere in the paper, exit probabilities are indexed from 0, so $p_t$ here is $p_{t-1}$ there.}

Training minimizes the exit-weighted language-modeling loss with entropy
regularization,
\begin{equation}
\mathcal{L}
=
\sum_{t=1}^{T} p_t\,\mathcal{L}_t
-
\beta H(p),
\end{equation}
using $\beta{=}0.15$ \rev{(Table~\ref{tab:looplm-recurrence})}. Here, $\mathcal{L}_t$ is the next-token
cross-entropy produced by the shared LM head from $h_t$. During pretraining,
loss is applied to all tokens; during mid-training, it is restricted to response
tokens. \rev{For the uniform distribution $U$ over the $T$ exits, $-\beta H(p)=\beta\,\mathrm{KL}(p\,\|\,U)-\beta\log T$, so $\beta$ is the KL coefficient of the main text.}

\begin{table}
\centering
\caption{Recurrence, exit gate, and training objective. The configuration is
shared by both training stages.}
\label{tab:looplm-recurrence}
\footnotesize
\setlength{\tabcolsep}{4pt}
\renewcommand{\arraystretch}{1.05}
\begin{tabular}{@{}p{0.34\linewidth}p{0.59\linewidth}@{}}
\toprule
\textbf{Parameter} & \textbf{Value} \\
\midrule
Recurrent steps $T$ & $4$ applications of the shared decoder \\
Initial state & Token embeddings $e$ \\
Later-step input & Previous hidden state $h_{t-1}$ \\
Embedding re-injection & None \\
Exit gate & Shared linear projection $2048\to1$ with sigmoid \\
Exit distribution & Learned per token; final step receives residual mass \\
LM objective & $\sum_t p_t \mathcal{L}_t$ \\
\rev{KL} coefficient $\beta$ & $0.15$ \\
Loss chunk size & $256$ positions \\
\bottomrule
\end{tabular}
\end{table}

\subsection{Pretraining}

Pretraining is a single from-scratch run of $100{,}000$ optimizer steps with
$2{,}097{,}152$ tokens per step, for a total of approximately $210$B tokens
(Table~\ref{tab:our-recipe}).

Training uses 64 GPUs with FSDP
(\texttt{shard\_grad\_op}), bfloat16 compute, fp32 gradient reduction and
buffers, FlashAttention-3, and activation checkpointing. Each GPU processes a
microbatch of 8 sequences of length $4096$. This already gives the full
2M-token optimizer batch, so gradient accumulation is one.

\rev{The pretraining corpus contains about 2.28T tokens, so the 210B-token
budget uses about 9\% of one pass and repeats no data.} The
mixture is dominated by filtered web text, with additional mathematics, code,
and instruction-following data.

\begin{table}
\centering
\caption{Pretraining mixture using mosaic-random sampling.}
\label{tab:looplm-pretrain-mix}
\small
\setlength{\tabcolsep}{7pt}
\renewcommand{\arraystretch}{1.05}
\begin{tabular}{@{}lr@{}}
\toprule
\textbf{Source} & \textbf{Weight} \\
\midrule
Nemotron-CC (high quality) & $66.5\%$ \\
Nemotron-CC-Math v1 & $15.0\%$ \\
MegaMath (high quality) & $4.6\%$ \\
Nemotron Cascade SFT Stage 1 & $9.6\%$ \\
Nemotron Post-Training (code) & $3.8\%$ \\
OpenCoder annealing corpus & $0.5\%$ \\
\bottomrule
\end{tabular}
\end{table}

\subsection{High-quality mid-training}

Mid-training starts from the step-$100{,}000$ pretraining checkpoint and continues
for $47{,}684$ steps with the same 2M-token batch, adding approximately $100$B
tokens.
\rev{The mixture contains the train splits of several evaluated benchmarks,
among them GSM8K, MATH and DROP, and was decontaminated against their test
sets by removing documents with 13-gram overlap with any test example.
We found no train splits of MMLU, MMLU-Pro or CRUXEval in it.}

The architecture, recurrent depth, \rev{KL} coefficient, weight decay, and
optimizer are unchanged. The peak learning rate is reduced to
$1\times10^{-4}$ and cosine-decayed to $1\times10^{-5}$, with a 1K-step
warmup.

Sequences remain 4K tokens long, but are formatted as instruction/response
pairs and the loss is applied only to response tokens. We use the HRM-Text data
collection of \citet{wang2026hrmtext}, reweighted toward mathematics and
reasoning while retaining a 42\% knowledge-preserving portion.

Dense baselines receive the same mid-training stage with identical data, schedule,
and token budget.

\begin{table}
\centering
\caption{Mid-training mixture based on HRM-Text sources. The first block preserves
general knowledge; the second emphasizes mathematics and reasoning.}
\label{tab:looplm-anneal-mix}
\small
\setlength{\tabcolsep}{7pt}
\renewcommand{\arraystretch}{1.05}
\begin{tabular}{@{}lr@{}}
\toprule
\textbf{Source} & \textbf{Weight} \\
\midrule
FLAN & $18.0\%$ \\
TaskSource & $9.0\%$ \\
Natural Reasoning & $7.0\%$ \\
WebInstruct-verified & $4.0\%$ \\
Platypus & $2.0\%$ \\
No Robots & $2.0\%$ \\
\midrule
SYNTH & $13.0\%$ \\
OpenMathInstruct-2 & $12.0\%$ \\
DeepMind Mathematics & $8.0\%$ \\
AceReason & $5.5\%$ \\
OpenThoughts-2 & $5.0\%$ \\
TextbookReasoning & $3.5\%$ \\
AMPS Mathematica & $3.0\%$ \\
NuminaMath & $2.5\%$ \\
Principia collection & $2.0\%$ \\
Sudoku-Extreme & $2.0\%$ \\
AMPS Khan & $0.8\%$ \\
MATH train / OmniMath / GSM8K train
& $0.3\%$ / $0.3\%$ / $0.1\%$ \\
\bottomrule
\end{tabular}
\end{table}

\subsection{Dense baselines and evaluation}

We use two dense counterparts to separate parameter-matched and
inference-compute-matched comparisons.

\texttt{dense-ouro-1.4B} uses the same 24-layer decoder block but removes
recurrence by setting $T{=}1$. It therefore matches LoopLM in parameter count.

\texttt{dense-ouro-3.9B} stacks three copies of the 24-layer block into a
72-layer dense model. Its inference cost matches LoopLM evaluated
at three recurrent steps.

Both dense models go through the same two training stages as LoopLM.

\rev{LoopLM is evaluated with a fixed exit after $k$ recurrent steps.
The inference-compute-matched comparison against \texttt{dense-ouro-3.9B}
uses $k{=}3$. Unless stated otherwise, the comparison with
\texttt{dense-ouro-1.4B}, including the gains quoted in the abstract, uses
$k{=}4$, as do the GSM8K and MMLU scores after mid-training in
Section~\ref{sec:looplm-final-setup}.} The four-step results in
Table~\ref{tab:looplm-benches-anneal} use additional, unmatched recurrent
compute. The learned exit gate is used during training, but fixed-depth
evaluation is used for the reported controlled comparisons.

\rev{Table~\ref{tab:eval-protocol} lists the evaluation settings of
Table~\ref{tab:looplm-benches-anneal}. All models are evaluated with the same
fork of lm-eval (version 0.4.9.1) on the same tasks, with LoopLM at a fixed exit
after $k$ iterations, a batch of 8 and a context of 4,096 tokens.}

\begin{table}[h]
\revcolor
\centering
\caption{Evaluation settings of Table~\ref{tab:looplm-benches-anneal}. The chat
format is the prompt format of the HRM-Text mid-training data.}
\label{tab:eval-protocol}
\small
\begin{tabular}{lccl}
\toprule
Benchmark & Shots & Format & Metric \\
\midrule
GSM8K & 0 & chat format, greedy generation & exact match of the boxed answer \\
MATH & 0 & chat format, greedy generation & math-verify \\
MATH500 & 0 & subset of the MATH run & math-verify \\
DROP & 3 fixed & chat format, greedy generation & F1 \\
MMLU & 5 & log-likelihood & accuracy \\
MMLU-Pro & 5 & chain of thought, at most 512 tokens & exact match \\
ARC-Challenge, ARC-Easy & 25 & log-likelihood & normalized accuracy \\
HellaSwag & 10 & log-likelihood & normalized accuracy \\
Winogrande & 5 & log-likelihood & accuracy \\
CRUXEval-O, CRUXEval-I & 2 & 10 samples, temperature 0.2 & pass@1 \\
\bottomrule
\end{tabular}
\end{table}

Reinforcement-learning settings are reported separately in
Appendix~\ref{app:rl}.

\section{Embedding injections}
\label{app:embedding-injections}

We also tried re-injecting the original token embeddings at every recurrent step: concatenate the initial embedding with the current state, then project back to hidden size before the next loop, following the input-injection construction of Universal Transformers \citep{dehghani2019universal} and later recurrent-depth models \citep{geiping2025scaling}. We crossed two choices for the injection's initial state $s_0$ (a learned zero vector vs.\ a truncated-normal draw) with two projection schemes (one shared projection vs.\ one per step), and compared all four against an otherwise identical injection-free run. The comparison is on training LM loss; none of these runs was evaluated downstream.

Table~\ref{tab:embedding-injection} reports the loss at a matched budget of 40K steps (2M-token batch, $\beta{=}0.15$, constant learning rate, no warmup). The four injection variants sit in a band of width $0.01$ around $1.90$--$1.91$, which is $0.12$--$0.13$ above the injection-free baseline ($1.78$). A per-step projection with a truncated-normal $s_0$ is the lowest of the four, by $0.005$, which we do not treat as a real difference. Concatenation injection therefore did not help in this \rev{setting}, and we keep the injection-free formulation.

\begin{table}[t]
\caption{Training LM loss with and without concatenation embedding injection, at a matched budget of 40K steps.}
\label{tab:embedding-injection}
\begin{center}
\small
\begin{tabular}{llcc}
\toprule
Projection & $s_0$ & Loss & $\Delta$ vs.\ none \\
\midrule
None (baseline) & --- & $1.78$ & --- \\
Per-step & truncated-normal & $1.90$ & $+0.12$ \\
Shared & truncated-normal & $1.91$ & $+0.12$ \\
Shared & learned zero & $1.91$ & $+0.12$ \\
Per-step & learned zero & $1.91$ & $+0.13$ \\
\bottomrule
\end{tabular}
\end{center}
\end{table}

\section{Looped dense architecture and training setups}
\label{app:looped-dense-setups}

\begin{figure}[!htbp]
\centering
\includegraphics[width=0.95\linewidth]{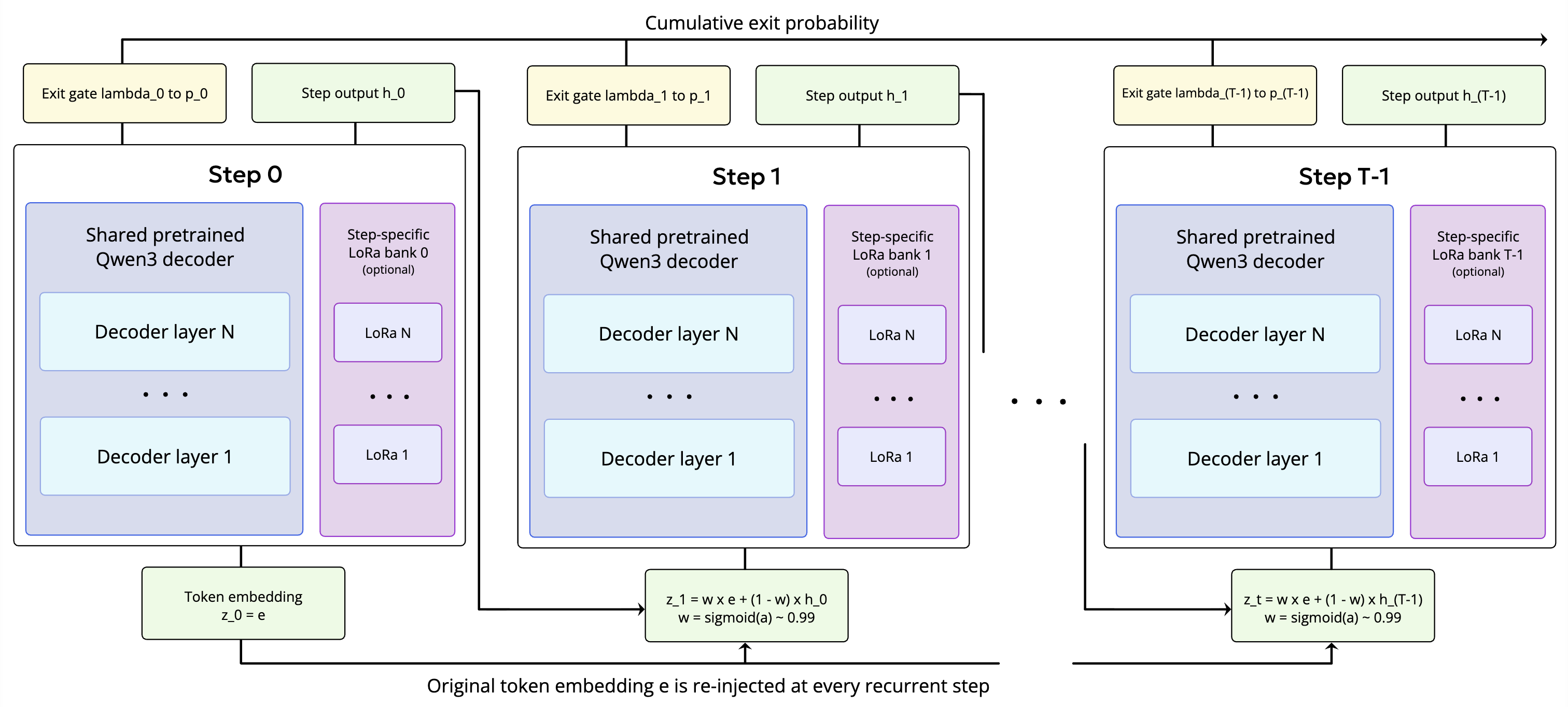}
\caption{Final Looped Qwen architecture. The pretrained decoder is shared across steps; optional LoRA adapters are step-specific. Later steps mix the original embeddings with the preceding hidden state, while a shared exit gate accumulates stopping probability.}
\label{fig:looped-dense-arch}
\end{figure}

Looped Qwen starts from Qwen3-1.7B-Base and applies the conversion procedure of
Section~\ref{sec:looped-dense}. The pretrained decoder, embeddings, \rev{and} vocabulary
are preserved and remain trainable. We add recurrent execution,
a shared exit gate, scalar embedding injection, \rev{an untied LM head initialized from the embeddings (311M parameters),} and, in one variant,
step-specific LoRA adapters.

The same 28-layer decoder is reused for $T{=}3$ recurrent steps, increasing
effective depth without duplicating the backbone. We run the conversion twice:
once on a curated instruction mixture for 60B tokens, and once on a web-scale
mixture for 84B tokens. Each run includes a fully shared model and a LoRA-32
variant, together with Qwen3-1.7B-Base and Qwen3-4B-Base baselines continued on
the same data for the same number of steps.

Unless stated otherwise, the configuration below refers to the 60B-token
instruction run with LoRA-32. The shared model differs only by removing the
step-specific LoRA banks; the web-scale run differs only in its data mixture
and training duration.

\subsection{Backbone}

The conversion keeps the architecture stored in the Qwen3-1.7B-Base checkpoint \rev{(Table~\ref{tab:looped-dense-backbone})}
rather than relying on fallback values from the generic model configuration.
All pretrained parameters remain trainable throughout continued training.

\begin{table}
\centering
\caption{Backbone configuration of the converted Qwen3-1.7B model.}
\label{tab:looped-dense-backbone}
\footnotesize
\setlength{\tabcolsep}{4pt}
\renewcommand{\arraystretch}{1.05}
\begin{tabular}{@{}p{0.34\linewidth}p{0.59\linewidth}@{}}
\toprule
\textbf{Parameter} & \textbf{Value} \\
\midrule
Initialization & Qwen3-1.7B-Base checkpoint \\
Trainable backbone & Decoder, embeddings, and LM head \\
Decoder layers & 28 \\
Hidden / FFN dimension & 2048 / 6144 \\
Attention heads & 16 query, 8 key--value \\
Attention head dimension & 128 \\
Attention pattern & Full causal GQA; no sliding window \\
Activation & SiLU-gated feed-forward network \\
Normalization & RMSNorm, $\epsilon=10^{-6}$ \\
Attention dropout / bias & 0 / disabled \\
Vocabulary size & 151,936 \\
Maximum positions & 32,768 \\
RoPE & Base $10^6$; no scaling \\
Input/output embeddings & Untied; LM head shared across recurrent steps \\
Initialization std. & 0.02 for newly added standard linear modules \\
\bottomrule
\end{tabular}
\end{table}

\subsection{Recurrence and input injection}

The pretrained decoder is applied recurrently for $T{=}3$ steps \rev{(Table~\ref{tab:looped-dense-recurrence})}.
The first step receives the original token embeddings $e$. For later steps,
we combine the original embeddings with the hidden state from the preceding
iteration:
\begin{equation}
z_t = w e + (1-w)h_{t-1},
\qquad t>0.
\end{equation}

The mixing coefficient is shared across all recurrent steps and parameterized
as
\begin{equation}
w=\sigma(a).
\end{equation}
We initialize $a=4.6$, corresponding to $w=0.99005$. Thus, later recurrent
steps remain anchored to the original token representation while incorporating
a small update from the previous hidden state.

A shared linear exit gate maps the 2048-dimensional hidden state to one scalar
per token. The model is trained with the same exit-weighted objective as the
from-scratch LoopLM, with \rev{KL} coefficient $\beta=0.15$. During conversion,
we additionally smooth the exit-weighted LM loss with $\epsilon=0.05$, ensuring
that every recurrent step receives a minimum direct LM weight of $0.0167$.

\begin{table}
\centering
\caption{Recurrent modules and objective used for Looped Qwen.}
\label{tab:looped-dense-recurrence}
\footnotesize
\setlength{\tabcolsep}{4pt}
\renewcommand{\arraystretch}{1.05}
\begin{tabular}{@{}p{0.34\linewidth}p{0.59\linewidth}@{}}
\toprule
\textbf{Parameter} & \textbf{Value} \\
\midrule
\multicolumn{2}{@{}l}{\textbf{Recurrence}} \\
Recurrent steps & $T=3$ applications of the shared decoder \\
Step 0 input & Original token embeddings $e$ \\
Later-step input & $z_t=w e+(1-w)h_{t-1}$ \\
Injection parameter & One trainable scalar shared across steps \\
Initialization & $a=4.6$, $w=\sigma(a)=0.99005$ \\
Exit gate & Shared linear projection $2048\rightarrow1$ \\
Exit mode & Learned per-token distribution \\
\midrule
\multicolumn{2}{@{}l}{\textbf{Objective}} \\
LM objective & Exit-weighted next-token cross-entropy \\
\rev{KL} coefficient $\beta$ & 0.15 \\
Exit smoothing $\epsilon$ & 0.05 \\
Minimum direct LM weight & 0.0167 per step \\
Target shift & 0; identical next-token target at every step \\
Loss mask / chunk size & Precomputed positional mask / 256 positions \\
\bottomrule
\end{tabular}
\end{table}

\subsection{Step-specific LoRA}

\rev{Apart from its untied output head, the shared variant adds only the scalar
input injection and the exit gate, and both are shared across recurrent steps.} The LoRA-32 variant additionally assigns an
independent LoRA bank to each recurrent step.

Adapters are applied to the attention $q$, $k$, $v$, and output projections and
to the MLP gate, up, and down projections. Each adapter has rank 32 and uses an
unscaled $BAx$ parameterization, with Kaiming-uniform initialization for $A$
and zero initialization for $B$.

Across three recurrent steps, the LoRA banks add 104,595,456 trainable
parameters, bringing the total number of trainable parameters in the LoRA-32
model to 2,136,339,458 \rev{(Table~\ref{tab:looped-dense-lora})}.

\begin{table}
\centering
\caption{Step-specific LoRA configuration.}
\label{tab:looped-dense-lora}
\footnotesize
\setlength{\tabcolsep}{4pt}
\renewcommand{\arraystretch}{1.05}
\begin{tabular}{@{}p{0.34\linewidth}p{0.59\linewidth}@{}}
\toprule
\textbf{Parameter} & \textbf{Value} \\
\midrule
Adapter layout & One independent bank per recurrent step \\
Rank & 32 \\
Attention targets & $q$, $k$, $v$, output projections \\
MLP targets & Gate, up, down projections \\
Parameterization & Unscaled $BAx$ \\
Initialization & Kaiming-uniform $A$, zero $B$ \\
LoRA parameters & 104,595,456 across three banks \\
Total trainable parameters & 2,136,339,458 \\
\bottomrule
\end{tabular}
\end{table}

\subsection{Continued-training setup}

The main conversion run uses the curated instruction mixture for 60B tokens,
corresponding to 28,611 optimizer steps \rev{(Table~\ref{tab:looped-dense-optimization})}. Sequence length is 4096.

Training uses 32 H100 GPUs across four nodes. Each GPU processes 16 sequences,
or 65,536 tokens. One distributed microbatch therefore contains 512 sequences,
corresponding to 2,097,152 tokens. This is already the full optimizer batch, so
gradient accumulation is one.

We use fused AdamW with
$\beta_1=0.9$, $\beta_2=0.95$, and $\epsilon=10^{-8}$. The learning rate
warms up linearly for 1,000 steps from $3\times10^{-6}$ to
$3\times10^{-5}$, then follows cosine decay to $3\times10^{-6}$.
Weight decay is 0.1 for attention and MLP projections and for LoRA parameters,
and zero otherwise. Gradients are clipped to global norm 1.0.

\begin{table}
\centering
\caption{Optimization and batch configuration for the 60B-token conversion run.}
\label{tab:looped-dense-optimization}
\footnotesize
\setlength{\tabcolsep}{4pt}
\renewcommand{\arraystretch}{1.05}
\begin{tabular}{@{}p{0.34\linewidth}p{0.59\linewidth}@{}}
\toprule
\textbf{Parameter} & \textbf{Value} \\
\midrule
Token budget & 60B tokens \\
Optimizer updates & 28,611 \\
Sequence length & 4096 \\
Optimizer & Fused AdamW \\
AdamW coefficients & $\beta_1=0.9$, $\beta_2=0.95$, $\epsilon=10^{-8}$ \\
Peak / minimum LR & $3\times10^{-5}$ / $3\times10^{-6}$ \\
LR schedule & 1K-step warmup, then cosine decay \\
Weight decay & 0.1 on attention/MLP projections and LoRA \\
Gradient clipping & Global norm 1.0 \\
\midrule
Hardware & 4 nodes $\times$ 8 H100 GPUs \\
Microbatch per GPU & 16 sequences / 65,536 tokens \\
Distributed microbatch & 512 sequences / 2,097,152 tokens \\
Gradient accumulation & 1 \\
Effective optimizer batch & 2,097,152 tokens \\
\bottomrule
\end{tabular}
\end{table}

\subsection{Training data}

The 60B-token run uses a six-source curated mixture dominated by mathematics,
code, and post-training data \rev{(Table~\ref{tab:looped-dense-mix})}. Sampling uses mosaic-random mixing with balanced
streams and shuffled random batches.

\begin{table}
\centering
\caption{Curated instruction mixture used for the 60B-token Looped Qwen run.}
\label{tab:looped-dense-mix}
\small
\setlength{\tabcolsep}{7pt}
\renewcommand{\arraystretch}{1.05}
\begin{tabular}{@{}p{0.68\linewidth}r@{}}
\toprule
\textbf{Source} & \textbf{Weight (\%)} \\
\midrule
Nemotron Post-Training Dataset v1 & 30.00 \\
Nemotron Post-Training Dataset v1 (math) & 27.90 \\
Nemotron Post-Training Dataset v1 (code) & 23.10 \\
Nemotron SFT Instruction Following Chat v2 & 11.90 \\
Nemotron Instruction Following Chat v1 & 3.86 \\
Nemotron Science v1 & 3.24 \\
\bottomrule
\end{tabular}
\end{table}

For reproducibility, training uses random seed 42, model-initialization seed
271,828, and shuffle seed 45. Shuffling is enabled with a block size of
10,737,418 samples.

\subsection{Baselines and evaluation}

Each conversion run is paired with two dense baselines.
\rev{Qwen3-1.7B-Base, the checkpoint from which Looped Qwen is initialized, is
the source-model reference; it keeps its tied output head, while Looped Qwen
trains an untied one.} Qwen3-4B-Base provides a larger,
inference-compute-matched reference. Both dense models continue training on exactly the
same mixture for the same number of optimizer steps as their corresponding
looped model. \rev{They use random seed 44 and shuffle seed 44 instead of 42 and 45, so their data order differs from that of Looped Qwen.}

The 60B-token instruction run is trained for 28,611 steps; the web-scale run
uses 84B tokens and 40,000 steps. In both cases, reported results use the final
checkpoint.

Looped Qwen is evaluated at a fixed exit after two of its three recurrent
steps, the comparison point used in Section~\ref{sec:looped-dense-results}.
\rev{All models in Table~\ref{tab:looped-dense-instruct} are evaluated with lm-eval, with the few-shot examples given as chat turns: GSM8K 8-shot (strict match), MATH 4-shot (Minerva), MMLU and MMLU-Pro 5-shot, ARC 25-shot, HellaSwag 10-shot, Winogrande 5-shot, BoolQ 0-shot, DROP 3-shot (F1) and CRUXEval-O 2-shot (pass@1).}

The learned gate is part of training, whereas the controlled
benchmark comparison uses a fixed recurrent depth.

\begingroup\revcolor
\paragraph{One recurrent step.}
The untied output head of Looped Qwen adds 311M parameters over its source model.
To check whether this head, rather than recurrence, explains the gains in Table~\ref{tab:looped-dense-instruct}, we evaluate the shared model of the instruction run after one recurrent step, with the same weights and protocol, except for the GSM8K few-shot examples (Table~\ref{tab:looped-dense-one-step}).
After one step it is level with or below Qwen3-1.7B-Base$^\dagger$ on seven of nine benchmarks.
It is above the baseline on GSM8K by 1.7 points and on MMLU by 0.2, both within one standard error of the difference.
The second step raises all nine scores. When both depths see the same prompts, it adds 2.4 points on GSM8K (McNemar $p\approx0.01$) and 2.4 F1 on DROP (95\% CI 1.9 to 2.9). On CRUXEval-O the paired gain of 0.2 is within noise (95\% CI $-1.1$ to $1.6$).
This suggests that the extra head does not explain the gains by itself.
The model was trained to run three steps, so one step is not a dense control with an untied head, and that control remains to be trained.

\begin{table}[!htbp]
\revcolor
\centering
\caption{\rev{Looped Qwen, shared, on the instruction mixture after one and after two of its three recurrent steps. The two-step column and the baseline repeat Table~\ref{tab:looped-dense-instruct}; standard errors are close to those given there. A two-step rerun on one GPU reproduced Table~\ref{tab:looped-dense-instruct} exactly on ARC-Challenge, ARC-Easy, MMLU, HellaSwag, Winogrande and BoolQ, so on these six benchmarks the one-step scores use the same prompts as that table. On GSM8K and DROP the one-step run drew different few-shot examples, and with the same examples two steps score 70.74 and 29.45 F1. On CRUXEval-O the prompts are the same, but generation is sampled, and a two-step rerun scores 29.52. MATH and MMLU-Pro were not evaluated after one step.}}
\label{tab:looped-dense-one-step}
\small
\begin{tabular}{lccc}
\toprule
Benchmark & Qwen3-1.7B-Base$^\dagger$ & One step & Two steps \\
\midrule
GSM8K & 66.64 & 68.31 & 70.96 \\
MMLU & 60.62 & 60.82 & 61.05 \\
ARC-Challenge & 53.84 & 53.33 & 55.46 \\
ARC-Easy & 79.97 & 79.25 & 80.13 \\
HellaSwag & 63.35 & 63.37 & 64.13 \\
Winogrande & 62.90 & 62.83 & 63.61 \\
BoolQ & 80.28 & 78.26 & 79.08 \\
DROP (F1) & 28.85 & 27.03 & 29.17 \\
CRUXEval-O & 30.48 & 29.30 & 29.90 \\
\bottomrule
\end{tabular}
\end{table}
\endgroup

\section{Conversion training dynamics}
\label{app:looped-dense-dynamics}

\begin{figure}[h]
\centering
\includegraphics[width=\textwidth]{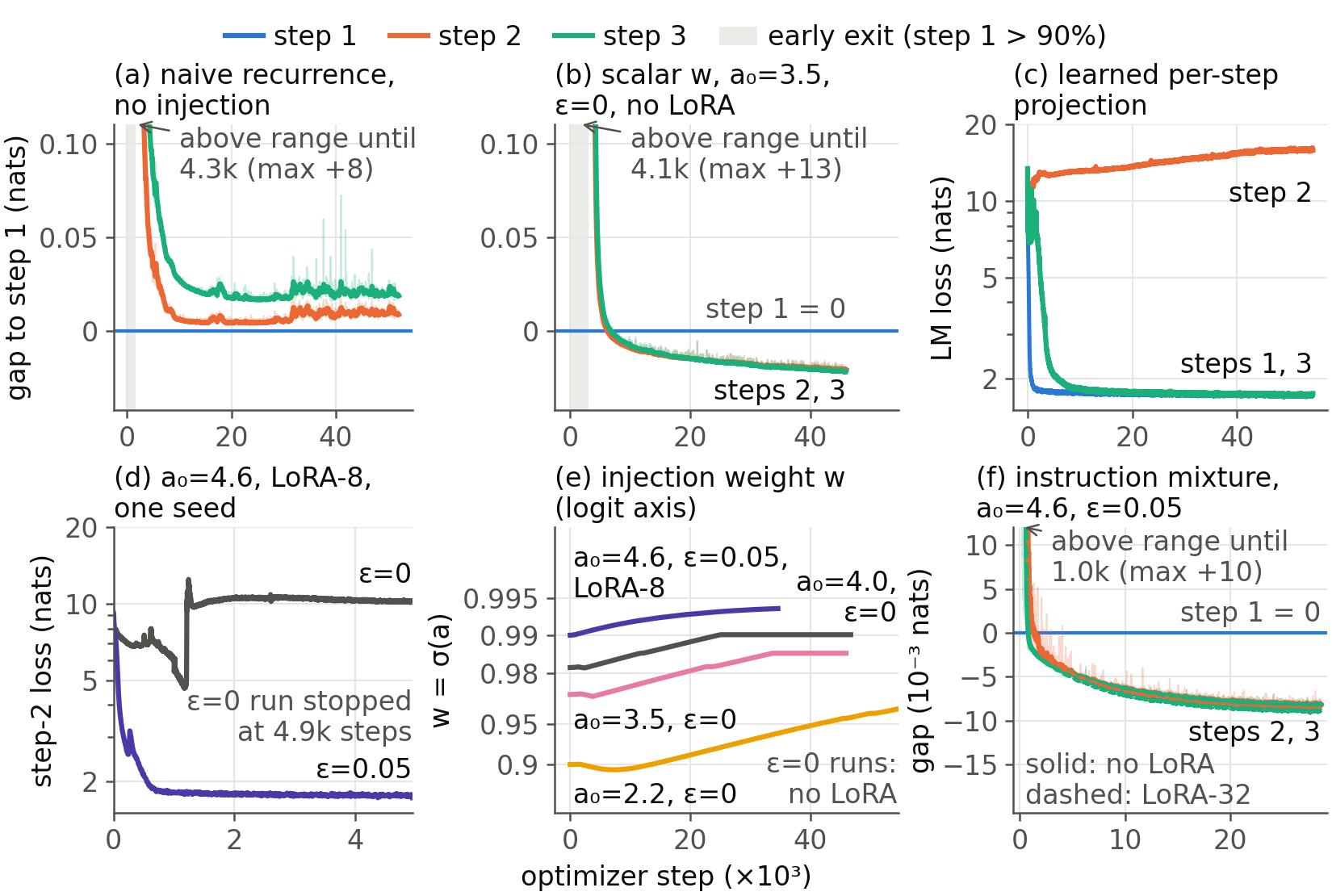}
\caption{Conversion training logs, one seed per configuration. \rev{Here, as in the text of Section~\ref{sec:looped-dense}, steps are counted from 1; the equations and Figure~\ref{fig:looped-dense-arch} count them from 0.}
(a) Naive recurrence without input re-injection and (b) scalar injection $w=\sigma(a)$ with $a_0=3.5$, $\epsilon=0$ and no LoRA: loss of steps 2 and 3 minus the loss of step 1 on the same batch. Shading marks where the exit probability of step 1 is above 0.9; before about 4k steps the gap is above the plotted range.
(c) Per-step LM loss of the run with a learned per-step projection.
(d) Step-2 loss at $a_0=4.6$ with rank-8 LoRA, without and with exit-loss smoothing; the unsmoothed run was stopped at 4.9k steps.
(e) Learned injection weight $w$ on a logit axis for $a_0=2.2$, $3.5$ and $4.0$ ($\epsilon=0$, no LoRA) and $a_0=4.6$ ($\epsilon=0.05$, rank-8 LoRA).
(f) The gap of (a) and (b) for the instruction-mixture runs with $a_0=4.6$ and $\epsilon=0.05$, without adapters (solid) and with rank-32 LoRA (dashed), in thousandths of a nat.
Plotted points are logged means over 5 to 10 optimizer steps (every step in the unsmoothed run of (d)); in (a), (b) and (f) the thick line is a 31-point moving average of these points.}
\label{fig:looped-dense-dynamics}
\end{figure}

Figure~\ref{fig:looped-dense-dynamics} shows the training logs behind Section~\ref{sec:looped-dense}.
Panels (a) and (b) come from two web-mixture runs that share data, batch, learning rate, $\beta{=}0.15$ and seed and differ only in the input injection.
The other panels come from single runs whose settings are given in the caption, and every configuration was trained with one seed.

\section{Conversion on the web-scale mixture}
\label{app:looped-dense-web}

\rev{To test whether the gains depend on the curated conversion data}, we repeat the same conversion procedure on a second mixture dominated by
filtered web text, with smaller mathematics and code components. This run uses
84B tokens and 40,000 optimizer steps. \rev{The instruction run uses 60B tokens and 28,611 steps, so the two runs differ in budget as well as in data.}

As in the instruction-data experiment, the looped and dense models are trained
on identical data for the same number of steps and evaluated at a fixed exit
after two recurrent iterations.

\begin{table}
\centering
\caption{
Conversion on the 84B-token web-scale mixture. Looped models are evaluated
after two of three recurrent steps. $\pm$ denotes standard error; $^\dagger$continued on the same
mixture for the same number of steps; bold marks scores above Qwen3-1.7B-Base$^\dagger$.
}
\label{tab:looped-dense-web}
\small
\setlength{\tabcolsep}{3pt}
\renewcommand{\arraystretch}{1.05}
\resizebox{\linewidth}{!}{
\begin{tabular}{lcccc}
\toprule
\textbf{Benchmark} &
\textbf{Qwen3-1.7B-Base}$^\dagger$ &
\textbf{Looped Qwen, shared} &
\textbf{Looped Qwen, LoRA-32} &
\textbf{Qwen3-4B-Base}$^\dagger$ \\
\midrule
GSM8K
& $60.12 \pm 1.35$
& $\mathbf{61.87 \pm 1.34}$
& $\mathbf{63.68 \pm 1.32}$
& $76.50 \pm 1.17$ \\
MATH
& $38.00 \pm 0.65$
& $\mathbf{38.08 \pm 0.65}$
& $\mathbf{39.22 \pm 0.65}$
& $47.12 \pm 0.66$ \\
MMLU
& $61.18 \pm 0.40$
& $\mathbf{62.73 \pm 0.39}$
& $\mathbf{62.15 \pm 0.39}$
& $71.04 \pm 0.36$ \\
MMLU-Pro
& $32.43 \pm 0.42$
& $\mathbf{32.96 \pm 0.42}$
& $\mathbf{33.88 \pm 0.42}$
& $46.28 \pm 0.44$ \\
ARC-Challenge
& $55.29 \pm 1.45$
& $\mathbf{55.38 \pm 1.45}$
& $54.61 \pm 1.45$
& $61.35 \pm 1.40$ \\
ARC-Easy
& $80.81 \pm 0.81$
& $\mathbf{81.27 \pm 0.80}$
& $80.81 \pm 0.80$
& $85.48 \pm 0.72$ \\
HellaSwag
& $69.37 \pm 0.46$
& $\mathbf{70.91 \pm 0.45}$
& $\mathbf{71.17 \pm 0.45}$
& $76.47 \pm 0.42$ \\
Winogrande
& $66.54 \pm 1.33$
& $\mathbf{67.56 \pm 1.32}$
& $\mathbf{68.19 \pm 1.31}$
& $71.98 \pm 1.26$ \\
\bottomrule
\end{tabular}
}
\end{table}

The shared looped model improves over Qwen3-1.7B-Base continued on the same data on every benchmark \rev{(Table~\ref{tab:looped-dense-web})},
although the gains are modest compared with the curated instruction run.
The largest improvements are $+1.75$ points on GSM8K, $+1.55$ on MMLU, and
$+1.54$ on HellaSwag; the remaining gains are about one point or less.

The LoRA-32 variant follows the same overall pattern and does not consistently
improve over the fully shared model. Qwen3-4B-Base remains ahead across all
benchmarks, consistent with the instruction-data experiment.

The two conversion runs therefore show the same qualitative effect across
substantially different data mixtures: recurrence improves over Qwen3-1.7B-Base given the same
continued training, while the size of the gain \rev{may depend} strongly on the
continued-training data.

\section{Compute accounting}
\label{app:compute}

Table~\ref{tab:compute} counts the compute per token of every model in the controlled comparisons, and three points follow from it.

The from-scratch recipe is cheap. The full two-stage run costs $9.9\times10^{21}$ training FLOPs. The same architecture trained on Ouro's schedule, 3T tokens at eight recurrent steps followed by 4.7T tokens at four, costs about $3.4\times10^{23}$, so our recipe needs about 3\% of that training compute, 35 times less.

The inference-compute match of Section~\ref{sec:baselines} is exact: LoopLM at three iterations and \texttt{dense-ouro-3.9B} apply the same 72 layer passes per token. Training LoopLM with all four iterations costs 1.4 times as much per token as \texttt{dense-ouro-3.9B}, which stores 2.7 times as many parameters.

Conversion is cheaper still. Converting Qwen3-1.7B-Base on the 60B-token mixture costs $1.9\times10^{21}$ FLOPs, 0.5\% of the $3.7\times10^{23}$ of its 36T-token pretraining. At its two-step exit, Looped Qwen uses 6.3 GFLOPs per token, 22\% fewer than Qwen3-4B-Base. The conversion keeps an untied LM head initialized from the input embeddings, and the counts include it.

\begin{table}[h]
\centering
\caption{Compute per token, with the usual approximation of $2N$ FLOPs for a forward pass and $6N$ for a training step, where $N$ counts the parameters applied to a token: the non-embedding layers times the recurrent steps, plus the LM head. Looped models train with all recurrent steps and apply the LM head at every step. Attention-score FLOPs are left out. \rev{They are equal for LoopLM at $k{=}3$ and \texttt{dense-ouro-3.9B}, and lower for Looped Qwen at $k{=}2$ than for Qwen3-4B-Base.} The Ouro-1.4B total is an estimate from its reported stage budgets and recurrent depths on the same architecture.}
\label{tab:compute}
\small
\setlength{\tabcolsep}{5pt}
\begin{tabular}{lccc}
\toprule
Model & Inference & Training & Total training \\
 & (GFLOPs/token) & (GFLOPs/token) & (FLOPs) \\
\midrule
LoopLM, exit $k{=}3$ / $k{=}4$ & 7.60 / 10.07 & 32.0 & $9.9\times10^{21}$ \\
\texttt{dense-ouro-1.4B} & 2.67 & 8.0 & $2.5\times10^{21}$ \\
\texttt{dense-ouro-3.9B} & 7.60 & 22.8 & $7.1\times10^{21}$ \\
Ouro-1.4B schedule (estimate) & 10.07 & 32.0 / 64.0 & $3.4\times10^{23}$ \\
\midrule
Looped Qwen, exit $k{=}2$ / $k{=}3$ & 6.26 / 9.08 & 31.0 & $1.9\times10^{21}$ (60B) \\
Qwen3-1.7B-Base & 3.44 & 10.3 & $3.7\times10^{23}$ (36T) \\
Qwen3-4B-Base & 8.04 & 24.1 & -- \\
\bottomrule
\end{tabular}
\end{table}

\begingroup\revcolor
\paragraph{Inference memory.}
Table~\ref{tab:memory} counts the weights and the KV cache at a context of 4,096 tokens in bf16.
LoopLM stores keys and values for every layer at every iteration, with 16 key-value heads and no sharing between iterations.
At three iterations its cache is therefore the same size as that of \texttt{dense-ouro-3.9B}, which applies the same 72 layers, and the two models differ only in their weights.
The saving is large at small batch sizes and small at large ones.
Measured with vLLM 0.26 on one H100 (Table~\ref{tab:decoding}), the cache takes 576 KiB per token both for LoopLM at three iterations and for \texttt{dense-ouro-3.9B}, as computed, and the two models decode at the same speed at every batch size.
LoopLM fits 30 sequences of 4,096 tokens in the cache against 28 for the dense model, and an FP8 cache doubles both numbers.
\texttt{dense-ouro-1.4B}, with the same parameters as LoopLM, decodes about three times faster, in line with its compute per token.
LoopLM can also draft for itself: the same weights at one iteration propose four tokens and the full model verifies them, with no separately trained draft model.
With an FP8 cache, 3.1 of the 4 draft tokens are accepted on average, and decoding at four iterations is 1.65 times faster at batch 1 and 1.53 times at batch 8; at three iterations the speed-up is 1.45 and 1.41 times, because one draft iteration is then a third of the model rather than a quarter.
Accuracy on 8-shot GSM8K does not change (65.6 without and 65.9 with drafting at four iterations, $n{=}1319$).
Table~\ref{tab:spec} compares this with ordinary speculative decoding of \texttt{dense-ouro-3.9B}, where \texttt{dense-ouro-1.4B} is the draft model, in the same setting.

\begin{table}[h]
\revcolor
\centering
\caption{Speculative decoding on one H100 with vLLM 0.26 and an FP8 cache, on 8-shot GSM8K prompts with up to 256 generated tokens and four draft tokens per step. Throughput is in generated tokens per second on the first 64 questions; accuracy uses all 1,319 questions with flexible answer extraction. ``Per step'' is the number of tokens accepted per verification, including the verified one. ``Sequences'' is how many 4,096-token sequences fit in the cache. Weights are as loaded by vLLM, about 0.03 GiB above the bf16 sizes of Table~\ref{tab:memory}.}
\label{tab:spec}
\small
\setlength{\tabcolsep}{4pt}
\begin{tabular}{llccccccc}
\toprule
Model & Draft & Batch 1 & Batch 8 & Per step & Weights (GiB) & KV (KiB/token) & Sequences & GSM8K \\
\midrule
\texttt{dense-ouro-3.9B} & none & 168 & 639 & -- & 7.30 & 288 & 56.0 & 65.8 \\
\texttt{dense-ouro-3.9B} & \texttt{dense-ouro-1.4B} & 251 & 905 & 4.15 & 9.97 & 384 & 39.8 & 65.7 \\
LoopLM, $k{=}3$ & none & 169 & 649 & -- & 2.70 & 288 & 60.1 & 63.5 \\
LoopLM, $k{=}3$ & one iteration & 245 & 917 & 4.07 & 5.38 & 384 & 42.8 & 62.9 \\
LoopLM, $k{=}4$ & none & 128 & 479 & -- & 2.70 & 384 & 45.1 & 65.6 \\
LoopLM, $k{=}4$ & one iteration & 211 & 731 & 4.05 & 5.38 & 480 & 34.2 & 65.9 \\
\bottomrule
\end{tabular}
\end{table}

\begin{table}[h]
\revcolor
\centering
\caption{Inference memory in GiB at a context of 4,096 tokens, bf16 weights and cache, without activations or framework overhead.}
\label{tab:memory}
\small
\begin{tabular}{lccccc}
\toprule
Model & Weights & KV per sequence & Batch 1 & Batch 8 & Batch 32 \\
\midrule
\texttt{dense-ouro-1.4B} & 2.67 & 0.75 & 3.42 & 8.67 & 26.67 \\
LoopLM, $k{=}3$ & 2.67 & 2.25 & 4.92 & 20.67 & 74.67 \\
LoopLM, $k{=}4$ & 2.67 & 3.00 & 5.67 & 26.67 & 98.67 \\
\texttt{dense-ouro-3.9B} & 7.27 & 2.25 & 9.52 & 25.27 & 79.27 \\
\bottomrule
\end{tabular}
\end{table}

\begin{table}[h]
\revcolor
\centering
\caption{Decoding on one H100 80GB with vLLM 0.26 at a context of 4,096 tokens (1,024 prompt and 3,072 generated tokens), median of two runs after warm-up. ``Sequences'' is how many 4,096-token sequences fit in the cache at 90\% memory utilization. Throughput is in generated tokens per second and latency in milliseconds per token. $^\ast$More sequences than fit in the cache, so vLLM preempted some of them.}
\label{tab:decoding}
\small
\setlength{\tabcolsep}{4pt}
\begin{tabular}{llcccccc}
\toprule
 & & & \multicolumn{2}{c}{Batch 1} & \multicolumn{2}{c}{Batch 8} & Batch 32 \\
\cmidrule(lr){4-5}\cmidrule(lr){6-7}\cmidrule(lr){8-8}
Model & Cache & Sequences & tok/s & ms/token & tok/s & ms/token & tok/s \\
\midrule
\texttt{dense-ouro-1.4B} & bf16 & 90.2 & 499 & 2.00 & 2{,}512 & 3.19 & 4{,}504 \\
LoopLM, $k{=}3$ & bf16 & 30.1 & 173 & 5.78 & 862 & 9.28 & 1{,}488$^\ast$ \\
LoopLM, $k{=}4$ & bf16 & 22.5 & 129 & 7.78 & 650 & 12.30 & 1{,}014$^\ast$ \\
\texttt{dense-ouro-3.9B} & bf16 & 28.0 & 174 & 5.74 & 864 & 9.26 & 1{,}454$^\ast$ \\
\midrule
\texttt{dense-ouro-1.4B} & FP8 & 180.2 & 486 & 2.06 & 2{,}895 & 2.76 & 6{,}018 \\
LoopLM, $k{=}3$ & FP8 & 60.1 & 165 & 6.07 & 988 & 8.10 & 2{,}066 \\
LoopLM, $k{=}4$ & FP8 & 45.1 & 127 & 7.89 & 757 & 10.57 & 1{,}552 \\
\texttt{dense-ouro-3.9B} & FP8 & 56.0 & 172 & 5.81 & 1{,}004 & 7.97 & 2{,}063 \\
\bottomrule
\end{tabular}
\end{table}
\endgroup

\section{Think-at-Hard}
\label{app:tah}
\label{app:think-at-hard}
We also reproduced Think-at-Hard (TaH) \citep{fu2025thinkhard} on our training mixture. TaH uses a static oracle policy to identify tokens that a frozen reference model predicts incorrectly on its first pass. The backbone and deeper-step LoRA adapters are then trained to refine these hard tokens, after which a lightweight decider is trained to imitate the oracle policy at inference. In our reproduction, the repeated-step LM loss never fell below the first-step loss\rev{, so the deeper step did not refine the prediction, and downstream accuracy did not improve over the first step}. Thus, we could not reproduce the gains reported in the original work under our data and training setup.
Despite this negative result, correctness-directed routing is an appealing alternative to the cumulative exit probabilities used by LoopLM. A deeper step receives supervision specifically where the shallower step fails, rather than learning a stopping probability only through its contribution to the expected LM loss. This directly connects additional computation to errors that still need to be corrected and may provide a cleaner objective for token-level adaptive depth.\\
The binary top-1 oracle used for two-step TaH does not directly specify how many iterations a hard token needs. An exact correctness-based extension to three steps would require labels indicating whether each token is first solved at step~0, 1, or 2. If these labels are precomputed, step~1 must first be trained before its errors can be measured, followed by another labeling pass and another training run for the three-step model. Repeating this procedure up to depth $T$ would require a sequence of shallower models and dataset relabeling stages. The current TaH paper avoids this cost for TaH-3 by assigning depths through quantiles of a frozen reference model's loss. This is efficient, but the resulting depth is a proxy for difficulty rather than the first recurrent step that actually predicts the token correctly.\\
We propose generating the depth mask online from the looped model itself. During training, we execute the maximum active number of recurrent steps and compare the top-1 prediction at every step with the target token. For token position $j$, we select
\begin{equation}
d_j = \min\{t : \arg\max_v q_t(v\mid x_{\leq j}) = x_{j+1}\},
\label{eq:dynamic-tah-depth}
\end{equation}
where $q_t$ is the token distribution at recurrent step $t$, and apply its LM gradient only at depth $d_j$. If no step predicts the target correctly, we assign the token to the final active step. The discrete mask is treated as a stop-gradient routing decision. Unlike static oracle labels, it always reflects the current model and requires neither a separate reference model nor an offline relabeling pass.
Naively enabling all depths from the start can starve intermediate steps of supervision. In a newly initialized three-step model, step~1 will often remain incorrect, so many tokens are routed directly from step~0 to step~2; step~1 then receives too few updates to improve. We address this with a depth curriculum. For a three-step model, the first half of training activates only steps~0 and~1, assigning all tokens missed by both to step~1. Step~2 is introduced in the second half, after step~1 has learned a useful refinement. More generally, the maximum active depth is increased one step at a time during training. This curriculum is intended to combine TaH's correctness-directed supervision with more than two recurrent steps, while avoiding the intermediate-step collapse \rev{that a fully dynamic oracle may cause}. We leave a controlled evaluation of this training rule to future work.

\begingroup\revcolor
\section{The larger run}
\label{app:scaled}

\paragraph{Setup.}
\scaled{} has 1.86B parameters: 1.23B in the shared layers and 0.62B in the untied embedding and output matrices.
Table~\ref{tab:scaled-stages} lists its training stages.
Stage two lowers the peak learning rate to $10^{-4}$ and reduces the Russian share of the mixture from about 30\% to about 22\%.

\begin{table}[!htbp]
\revcolor
\caption{Training stages of \scaled. Budgets are per stage. The learning-rate warm-up lasts 1,000 steps in pretraining, 500 at 16K, 300 at 32K, 100 in SFT and 10 in RL.}
\label{tab:scaled-stages}
\begin{center}
\small
\begin{tabular}{llll}
\toprule
Stage & Context & Budget & Learning rate, notes \\
\midrule
Pretraining 1 & 4K & 100,000 steps, 419B tokens & $3\times10^{-4}$ to $3\times10^{-5}$, RoPE $4\times10^{4}$ \\
Pretraining 2 & 4K & 200,000 steps, 839B tokens & $10^{-4}$ to $10^{-5}$ \\
Context 16K & 16K & 47,684 steps, 200B tokens & $2\times10^{-5}$ to $2\times10^{-6}$, RoPE $10^{6}$, document masking \\
Context 32K & 32K & 4,769 steps, 20B tokens & $10^{-5}$ to $10^{-6}$, RoPE $4\times10^{6}$, document masking \\
SFT & 32K & 7,000 steps, 14.7B tokens & $10^{-5}$ to $10^{-6}$, document masking \\
RL & 16K & step 90 of 150 & GRPO on the final iteration, math and code, $10^{-6}$ \\
\bottomrule
\end{tabular}
\end{center}
\end{table}

\paragraph{Dense twins.}
Two dense models were trained alongside the first pretraining stage with the data, batch, learning-rate schedule, seed and untied head of \scaled{}, and no model uses EMA.
The 24-layer twin has the parameter count of \scaled{}, and the 72-layer twin (4.3B parameters) has its inference compute at three iterations.
The 24-layer twin was also continued through the first half of the second stage.
Table~\ref{tab:scaled-twins} evaluates these pretraining checkpoints in one setup.
LoopLM leads the 24-layer twin on all 11 benchmarks at 419B and at 839B tokens, on BoolQ at 839B tokens by only 0.1 points.
The 72-layer twin leads LoopLM on all 11 at 419B tokens, by 1.0 to 11.0 points.
This is the ordering of Table~\ref{tab:looplm-benches-anneal}, with a second tokenizer and mixture and without instruction-formatted mid-training.
The gaps to the 72-layer twin are larger than in Table~\ref{tab:looplm-benches-anneal}, 11.0 points on GSM8K and 7.4 on MATH, so the near-parity on mathematics there may owe part of its size to mid-training.

\begin{table}[!htbp]
\revcolor
\centering
\caption{Pretraining checkpoints of \scaled{} and its dense twins, after 419B tokens (end of the first stage) and 839B tokens (half of the second). LoopLM runs three iterations at 419B and four at 839B. Evaluation uses lm-eval without a chat template: ARC 25-shot (normalized accuracy), HellaSwag 10-shot (normalized accuracy), Winogrande 5-shot, BoolQ 0-shot, CRUXEval 2-shot pass@1, GSM8K 8-shot strict match, MATH 4-shot (Minerva), MMLU and MMLU-Pro 5-shot.}
\label{tab:scaled-twins}
\small
\begin{tabular}{lccccc}
\toprule
 & \multicolumn{3}{c}{419B tokens} & \multicolumn{2}{c}{839B tokens} \\
\cmidrule(lr){2-4}\cmidrule(lr){5-6}
Benchmark & LoopLM & 24 layers & 72 layers & LoopLM & 24 layers \\
\midrule
GSM8K & 33.81 & 16.98 & 44.81 & 45.79 & 25.40 \\
MATH & 20.56 & 13.94 & 27.92 & 30.42 & 19.80 \\
MMLU & 46.62 & 37.32 & 53.45 & 51.87 & 43.17 \\
MMLU-Pro & 18.18 & 11.69 & 24.33 & 25.55 & 19.13 \\
ARC-Challenge & 48.29 & 44.71 & 53.75 & 50.09 & 45.65 \\
ARC-Easy & 76.98 & 74.75 & 80.98 & 77.65 & 75.25 \\
HellaSwag & 66.68 & 61.39 & 71.99 & 66.56 & 60.91 \\
Winogrande & 62.90 & 58.88 & 65.82 & 63.46 & 60.14 \\
BoolQ & 64.34 & 61.62 & 70.15 & 67.03 & 66.91 \\
CRUXEval-I & 24.69 & 18.97 & 25.70 & 30.41 & 25.88 \\
CRUXEval-O & 18.91 & 14.46 & 20.90 & 20.71 & 14.26 \\
\bottomrule
\end{tabular}
\end{table}

\paragraph{Evaluation.}
All models in Table~\ref{tab:scaled-results} are evaluated in one harness.
Reasoning uses a 16K budget with forced closing, and 30K for GPQA and the olympiad tasks; function calling runs without forcing.
The models differ in how many parameters they apply per generated token, counting repeated layers each time.
\scaled{} has 1.86B parameters and applies 5.24B per token.
Ouro-1.4B-Thinking and Ouro-2.6B-Thinking have 1.43B and 2.67B and apply 5.03B and 9.97B.
Qwen3.5-2B and SmolLM3-3B are dense, with 1.88B and 3.08B.

\paragraph{Inference.}
A looped model can draft for itself: the same weights at one iteration propose tokens and the full model verifies them, with no separately trained draft model.
On an earlier SFT checkpoint of this run on one H200, with four draft tokens, 3.8 to 4.1 tokens were accepted per verification, including the verified one, as in Table~\ref{tab:spec}.
Together with an FP8 KV cache this decoded 1.73 times faster than bf16 at batch 1 and 16K context, 2.10 times at $16\times8$K and 2.24 times at $8\times16$K.
Accuracy on MATH-500 and GSM8K with reasoning did not change ($n{=}1819$).
FP8 alone holds 2.1 times as many tokens in the cache.

\paragraph{Sharing the KV cache.}
Every layer stores separate keys and values at every iteration.
Reusing the last iteration's cache for decoded tokens, which Ouro reports as close to lossless on its benchmarks, drifted on our long reasoning traces: the KL divergence from the full cache grew from 0.03 to 3.6 over 4,096 decoded tokens.
Collapsing all iterations into one cache failed outright.
Keeping the first iteration separate and sharing the other three gave twice the capacity with a KL divergence of $4.9\times10^{-3}$ that did not grow with length, and few-shot GSM8K fell by 3.7 points (50.7 to 47.0, $n{=}300$), \rev{a drop within sampling noise}.
\endgroup

\begingroup\revcolor
\section{Comparison with published models}
\label{app:external}

Because published models differ substantially in evaluation protocol and pretraining budget,
Table~\ref{tab:looplm-external} is a contextual rather than a controlled comparison.
\rev{LoopLM trails Ouro-1.4B, trained on 25$\times$ more tokens, on every benchmark except GSM8K.
The protocols of the published rows differ from ours on GSM8K, MATH500 and MMLU-Pro, among others, as the caption describes.}
Our mid-training data contains the train splits of several of these benchmarks, including
GSM8K, so these cross-model comparisons are contextual.

\begin{table}[!htbp]
\revcolor
\caption{LoopLM after mid-training against published models between 1B and 4B
parameters, on the benchmarks reported by \citet{zhu2025ouro}. Bold marks the best score in each column; underline marks the
second best. The GSM8K marks leave out LoopLM, whose 0-shot chat protocol is not comparable. All rows other than LoopLM and Huginn-3.5B are taken from Table~7 of
\citet{zhu2025ouro}, which evaluates them with one pipeline; Huginn-3.5B values are from
\citet{geiping2025scaling} at 32 recurrences. These protocols differ from ours
(Table~\ref{tab:eval-protocol}), for example 3-shot chain of thought for GSM8K and 5-shot chain
of thought for MATH500 and MMLU-Pro. Tokens are training tokens in trillions.}
\label{tab:looplm-external}
\begin{center}
\resizebox{\linewidth}{!}{%
\begin{tabular}{lrrrrrrrr}
\toprule
Model & Tokens & GSM8K & MATH500 & MMLU & MMLU-Pro & ARC-C & HellaSwag & Winogrande \\
\midrule
\textbf{LoopLM (ours), 4 iter.}
& 0.31
& 83.3
& 63.2
& 56.8
& 24.5
& 51.0
& 59.3
& 63.9 \\
\textbf{LoopLM (ours), 3 iter.}
& 0.31
& 81.6
& 63.6
& 55.9
& 24.3
& 50.5
& 58.9
& 62.9 \\
\midrule
Ouro-1.4B
& 7.7
& \textbf{78.9}
& \textbf{82.4}
& \underline{67.4}
& \underline{48.6}
& \underline{60.9}
& 74.3
& \textbf{72.3} \\
Huginn-3.5B
& 0.8
& 32.6
& ---
& 31.4
& ---
& 38.2
& 65.2
& 59.4 \\
Gemma-3-1B
& 2
& 2.1
& 41.0
& 39.9
& 11.3
& 39.3
& 56.1
& 58.7 \\
Llama-3.2-1B
& 9
& 7.1
& 7.4
& 45.5
& 11.8
& 42.0
& 59.4
& 62.8 \\
Qwen2.5-1.5B
& 18
& 60.7
& 17.6
& 61.0
& 29.1
& 54.4
& 67.7
& 66.8 \\
Qwen3-1.7B
& 36
& 70.3
& 25.8
& 62.5
& 37.3
& 55.7
& 67.1
& 66.3 \\
Qwen2.5-3B
& 18
& \underline{74.6}
& 42.6
& 65.6
& 37.9
& 55.5
& 74.5
& 70.2 \\
Llama-3.2-3B
& 9
& 67.2
& 40.8
& 59.7
& 33.3
& 52.5
& 73.1
& 69.1 \\
Gemma-3-4B
& 4
& 68.7
& \underline{68.6}
& 58.4
& 34.6
& \underline{60.9}
& \underline{75.6}
& 71.1 \\
Qwen3-4B
& 36
& 72.9
& 59.6
& \textbf{73.2}
& \textbf{51.4}
& \textbf{63.7}
& \textbf{75.7}
& \underline{71.2} \\
\bottomrule
\end{tabular}}
\end{center}
\end{table}
\endgroup

\begingroup\revcolor
\section{Held-out language modeling}
\label{app:lm}

Table~\ref{tab:lm} reports zero-shot LAMBADA accuracy and wikitext-2 bits per byte for the from-scratch checkpoints of this paper, the larger run of Section~\ref{sec:scaled} and Ouro-1.4B Base.
All models run in one setup: lm-eval 0.4.13 with vLLM, bf16, a 4,096-token context and one leading end-of-text token.
Bits per byte compare models with different tokenizers. The from-scratch models of this paper share the tokenizer of Ouro, and the larger run uses the Qwen3 tokenizer.

The larger run comes within 3.7 points of Ouro-1.4B Base on LAMBADA and within 0.036 bits per byte on wikitext, with 5.2 times fewer pretraining tokens.
Its pretraining checkpoints show the ordering of Table~\ref{tab:looplm-benches-anneal}.
LoopLM leads the dense twin with the same parameters, by 1.5 points on LAMBADA at 419B tokens and by 5.5 points at 839B, and it trails the twin with the same compute per token.
The from-scratch checkpoints of this paper end with the instruction-formatted mid-training stage.
After that stage LoopLM trails its parameter-matched twin on both measures, although both models went through it. Table~\ref{tab:lm} has no checkpoints of this paper from before mid-training, so it cannot show whether that stage is why the ordering differs from the larger run.

\begin{table}[!htbp]
\revcolor
\centering
\caption{Zero-shot LAMBADA accuracy and wikitext-2 bits per byte (lower is better). The twins of the larger run share its data, batch, learning-rate schedule and seed; 24 layers matches LoopLM in parameters and 72 layers matches it in compute per token at three iterations. At 419B tokens LoopLM scores the same at three and four iterations. At 419B and 839B tokens the two larger-run models have the same wikitext bits per byte to three decimals, while their word perplexities differ slightly: 14.69 and 14.71 for LoopLM, 15.77 and 15.78 for the 24-layer twin. Comparisons across the three groups are not controlled.}
\label{tab:lm}
\small
\begin{tabular}{llcc}
\toprule
Model & \rev{Training tokens} & LAMBADA & wikitext \\
\midrule
LoopLM, three iterations & 310B & 44.6 & 0.809 \\
LoopLM, four iterations & 310B & 39.9 & 0.810 \\
\texttt{dense-ouro-1.4B} & 310B & 50.2 & 0.796 \\
\texttt{dense-ouro-3.9B} & 310B & 55.1 & 0.736 \\
\midrule
Larger run, LoopLM & 419B & 57.7 & 0.725 \\
\quad dense, 24 layers & 419B & 56.2 & 0.744 \\
\quad dense, 72 layers & 419B & 65.4 & 0.683 \\
Larger run, LoopLM & 839B & 61.6 & 0.725 \\
\quad dense, 24 layers & 839B & 56.1 & 0.744 \\
\scaled{}, end of 4K pretraining & 1.26T & 61.9 & 0.711 \\
\scaled{}, after 16K and 32K context & 1.48T & 62.3 & 0.707 \\
\midrule
Ouro-1.4B Base & 7.7T & 66.0 & 0.671 \\
\bottomrule
\end{tabular}
\end{table}
\endgroup

\end{document}

%% file: math_commands.tex
\usepackage{amsmath,amsfonts,bm}

\def\eqref#1{equation~\ref{#1}}

\def\1{\bm{1}}

\DeclareMathAlphabet{\mathsfit}{\encodingdefault}{\sfdefault}{m}{sl}
\SetMathAlphabet{\mathsfit}{bold}{\encodingdefault}{\sfdefault}{bx}{n}

